\documentclass[11pt]{article}

\usepackage[final]{acl}

\usepackage{times}
\usepackage{latexsym}



\usepackage[T1]{fontenc}

\usepackage[utf8]{inputenc}

\usepackage{mathtools}

\usepackage{microtype}

\usepackage{inconsolata}

\usepackage{graphicx}

\usepackage{amsmath,amsfonts,bm}

\def\eqref#1{equation~\ref{#1}}

\def\1{\bm{1}}

\DeclareMathAlphabet{\mathsfit}{\encodingdefault}{\sfdefault}{m}{sl}
\SetMathAlphabet{\mathsfit}{bold}{\encodingdefault}{\sfdefault}{bx}{n}

\usepackage{url}
\usepackage{inconsolata}
\usepackage{times}
\usepackage{latexsym}
\usepackage{amsfonts}
\usepackage{amsmath}
\usepackage{amssymb}
\usepackage{multicol}
\usepackage{multirow}
\usepackage{xspace}
\usepackage{booktabs}
\usepackage{bbding}
\usepackage{array}
\usepackage{threeparttable}
\usepackage{tcolorbox}
\usepackage{tabularx}
\usepackage{enumitem}
\usepackage{xcolor,colortbl}
\usepackage{color}
\usepackage{setspace}
\usepackage{makecell}
\usepackage{listings}
\usepackage{xltabular}
\usepackage{tcolorbox}
\tcbuselibrary{breakable}
\usepackage{xcolor}

\lstdefinestyle{python}{
    language=Python,
    basicstyle=\ttfamily\small,
    keywordstyle=\color{blue}\bfseries,
    commentstyle=\color{green},
    stringstyle=\color{red},
    numberstyle=\tiny\color{gray},
    showstringspaces=false,
    frame=single,
    breaklines=true,
    backgroundcolor=\color{lightgray!20}
}

\usepackage{booktabs}

\usepackage[utf8]{inputenc} 
\usepackage[T1]{fontenc}    
\usepackage{hyperref}       
\usepackage{url}            
\usepackage{booktabs}       
\usepackage{amsfonts}       
\usepackage{nicefrac}       
\usepackage{microtype}      
\usepackage{xcolor}         
\usepackage{float}
\usepackage{graphicx}       
\usepackage{wrapfig}
\usepackage{subcaption}
\usepackage{makecell}
\usepackage{soul}

\hypersetup{
    colorlinks=true,
    linkcolor=blue,
}

\definecolor{Ocean}{RGB}{129,194,250}

\definecolor{deepgreen}{RGB}{0, 70, 0}
\usepackage{cleveref}
\crefformat{section}{\S#2#1#3}
\crefformat{subsection}{\S#2#1#3}
\crefformat{subsubsection}{\S#2#1#3}
\crefrangeformat{section}{\S\S#3#1#4 to~#5#2#6}
\crefmultiformat{section}{\S\S#2#1#3}{ and~#2#1#3}{, #2#1#3}{ and~#2#1#3}
\crefmultiformat{subsection}{\S\S#2#1#3}{ and~#2#1#3}{, #2#1#3}{ and~#2#1#3}
\Crefformat{figure}{#2Fig.~#1#3}
\Crefmultiformat{figure}{Figs.~#2#1#3}{ and~#2#1#3}{, #2#1#3}{ and~#2#1#3}
\Crefformat{table}{#2Tab.~#1#3}
\Crefmultiformat{table}{Tabs.~#2#1#3}{ and~#2#1#3}{, #2#1#3}{ and~#2#1#3}
\Crefformat{appendix}{Appx.~\S#2#1#3}
\crefmultiformat{appendix}{Appx.~\S#2#1#3}{ and~#2#1#3}{, #2#1#3}{ and~#2#1#3}
\crefformat{algorithm}{Alg.~#2#1#3}
\Crefformat{equation}{Eq.~#2#1#3}

\makeatother

\definecolor{myred}{rgb}{0.7, 0.3, 0.0}
\definecolor{myblue}{HTML}{054488}
\definecolor{mygreen}{HTML}{056b34}
\definecolor{myorange}{HTML}{ff8800}
\definecolor{mypurple}{HTML}{8400ff}
\definecolor{mypink}{HTML}{f7acb9}

\definecolor{myred}{rgb}{0.7, 0.3, 0.0}
\definecolor{myblue}{HTML}{054488}
\definecolor{mygreen}{HTML}{056b34}

\usepackage{times}
\usepackage{CJKutf8}
\usepackage{latexsym}
\usepackage{tcolorbox}
\usepackage{multirow}
\usepackage{wrapfig}
\usepackage{tikz}
\usepackage{capt-of}
\usepackage{graphicx}  
\usepackage{pgfplots}
\pgfplotsset{compat=1.12}
\usepackage{amsmath}
\usepackage{multicol}
\usepackage{color}
\usepackage{mwe}
\usepackage{wrapfig}
\usepackage{colortbl,array}
\usepackage{xspace}
\usepackage{tikz}
\usetikzlibrary{tikzmark}
\makeatletter
\newcommand*\myfontsize{%
  \@setfontsize\myfontsize{7}{8}%
}
\makeatother

\definecolor{myred}{rgb}{0.7, 0.3, 0.0}
\definecolor{myblue}{HTML}{054488}
\definecolor{mygreen}{HTML}{056b34}

\newcolumntype{R}[1]{>{\raggedleft\let\newline\\\arraybackslash\hspace{0pt}}m{#1}}

\usetikzlibrary{intersections}

\definecolor{darkgreen}{rgb}{0.0, 0.42, 0.24}
\definecolor{lightblue}{RGB}{221,235,247}
\usepackage{caption}
\usepackage{subcaption}
\usepackage{graphicx}
\usepackage{pifont}
\usepackage{titletoc}
\usepackage{amsfonts}
\usepackage{booktabs}
\usepackage{arydshln}
\usepackage{colortbl}
\usepackage{algorithm}
\usepackage[noend]{algpseudocode}
\usepackage{enumitem}
\usepackage{graphicx}
\usepackage{soul}
\usepackage{colortbl,array,xcolor}
\usepackage{amsmath}
\usepackage{fontawesome}

\title{Entropy-Adaptive Fine-Tuning: \\ Resolving Confident Conflicts to Mitigate Forgetting}

\author{
    Muxi Diao$^{1,2}${\hypersetup{linkcolor=black}\thanks{Equal contribution.}}, 
    Lele Yang$^1${\hypersetup{linkcolor=black}\footnotemark[1]}, 
    Wuxuan Gong$^1${\hypersetup{linkcolor=black}\footnotemark[1]}, 
    Yutong Zhang$^1$, \\
    \textbf{Zhonghao Yan$^1$}, 
    \textbf{Yufei Han$^1$}, 
    \textbf{Kongming Liang$^1$}, 
    \textbf{Weiran Xu$^1$}, 
    \textbf{Zhanyu Ma$^1${\hypersetup{linkcolor=black}\thanks{Corresponding author.}}} \\
    $^1$Beijing University of Posts and Telecommunications, $^2$Zhongguancun Academy\\
    \texttt{\{dmx, yang\_happy, mazhanyu\}@bupt.edu.cn}\\
    \\
    \begin{tabular}{@{}ll@{}}
    \faGithub\ GitHub: \href{https://github.com/PRIS-CV/EAFT}{\texttt{\textcolor[RGB]{119,171,243}{https://github.com/PRIS-CV/EAFT}}}
    \end{tabular}
}

\begin{document}
\maketitle


\begin{abstract}
Supervised Fine-Tuning (SFT) is the standard paradigm for domain adaptation, yet it frequently incurs the cost of catastrophic forgetting. In sharp contrast, on-policy Reinforcement Learning (RL) effectively preserves general capabilities. We investigate this discrepancy and identify a fundamental distributional gap: while RL aligns with the model's internal belief, SFT forces the model to fit external supervision. This mismatch often manifests as \textbf{"Confident Conflicts"}—tokens characterized by low probability but low entropy. In these instances, the model is highly confident in its own prediction but is forced to learn a divergent ground truth, triggering destructive gradient updates.
To address this, we propose \textbf{Entropy-Adaptive Fine-Tuning (EAFT)}. Unlike methods relying solely on prediction probability, EAFT utilizes token-level entropy as a gating mechanism to distinguish between epistemic uncertainty and knowledge conflict. This allows the model to learn from uncertain samples while suppressing gradients on conflicting data. Extensive experiments on Qwen and GLM series (ranging from 4B to 32B parameters) across mathematical, medical, and agentic domains confirm our hypothesis. EAFT consistently matches the downstream performance of standard SFT while significantly mitigating the degradation of general capabilities.
\end{abstract}
\section{Introduction}
Supervised Fine-Tuning (SFT) is the standard method for adapting Large Language Models (LLMs) to specific domains (e.g., mathematics or agentic tool-use)~\citep{qwenmath,shao2024deepseekmath,k2}. However, this paradigm often comes with a significant cost known as catastrophic forgetting \citep{kirkpatrick2017overcoming,ouyang2022training}. Previous studies have extensively documented that while fitting specific target distributions, models frequently suffer from degradation in general capabilities \citep{ouyang2022training,luo2023empirical}.
In contrast, on-policy Reinforcement Learning (RL) has shown a remarkable ability to improve domain-specific performance while effectively preserving the robustness of the base model \citep{chen2025retaining,rlRazor}. This sharp contrast raises a fundamental question: 

\begin{center}
\textbf{\textit{Why does SFT frequently degrade general abilities, while on-policy RL preserves them?}}
\end{center}

\begin{figure*}[t]
    \centering
    \includegraphics[width=\linewidth]{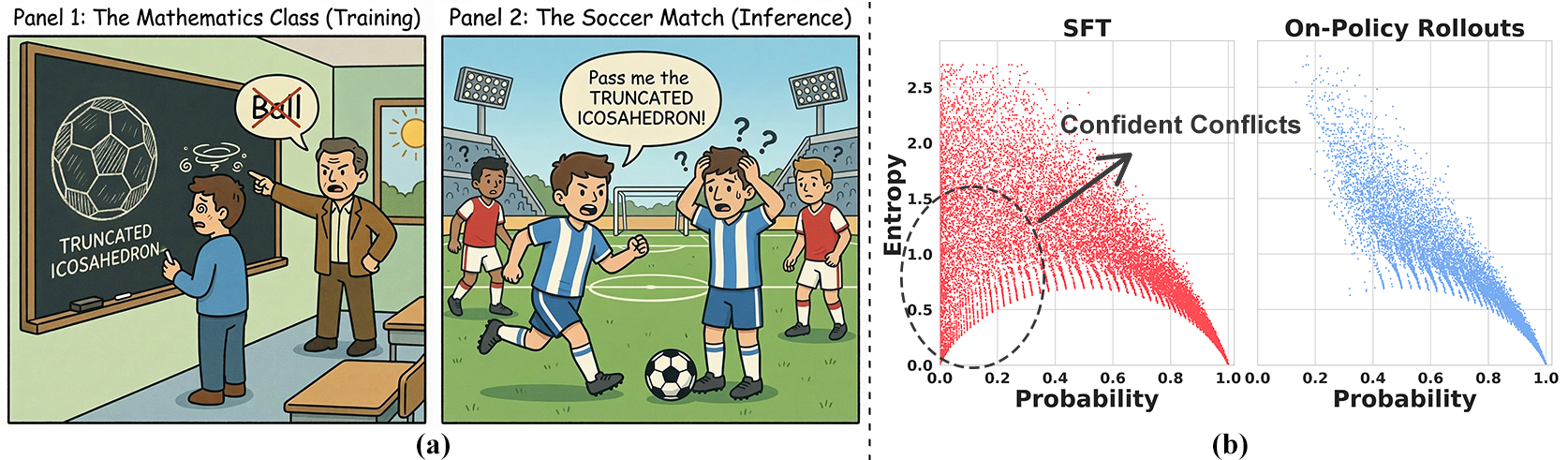}
    \caption{\textbf{(a)} Conceptual illustration. When SFT forces the model to override its strong priors (e.g., labeling a ``ball'' as a ``truncated icosahedron''), it creates a \textbf{\textit{Confident Conflict}}. 
Fitting these conflicts distorts the model's existing representations, leading to catastrophic forgetting. \textbf{(b)} Token-level entropy–probability landscape. Compared to on-policy rollouts (right), the SFT data (left) exhibits a prominent cluster of \textbf{\textit{Low Entropy, Low Probability}} tokens.}
    \label{fig:main}
\end{figure*}

To investigate the mechanisms behind this phenomenon, we systematically analyze the token-level probability and entropy of the training data. 
As visualized in Fig.~\ref{fig:main}, this analysis reveals a distinct distributional gap arising from different data sources. 
In on-policy RL, training sequences are generated via self-rollout; consequently, the tokens inherently align with the model's current probability landscape, falling into either high-probability confidence zones or high-entropy exploration regions. 
Conversely, SFT relies on external supervision (e.g., humans or strong teacher models), introducing a mismatch manifested as low-probability, low-entropy tokens. Crucially, this mismatch manifests as tokens characterized by \textbf{low probability yet low entropy}. 
This specific region corresponds to scenarios where the model is highly confident in its own prediction (low entropy) but is forced to fit a divergent ground-truth label (low probability). We term these instances \textbf{"Confident Conflicts"}. See App. \ref{app:entropy_analysis} for representative word clouds.

\begin{figure}[t]
    \centering\includegraphics[width=0.99\linewidth]{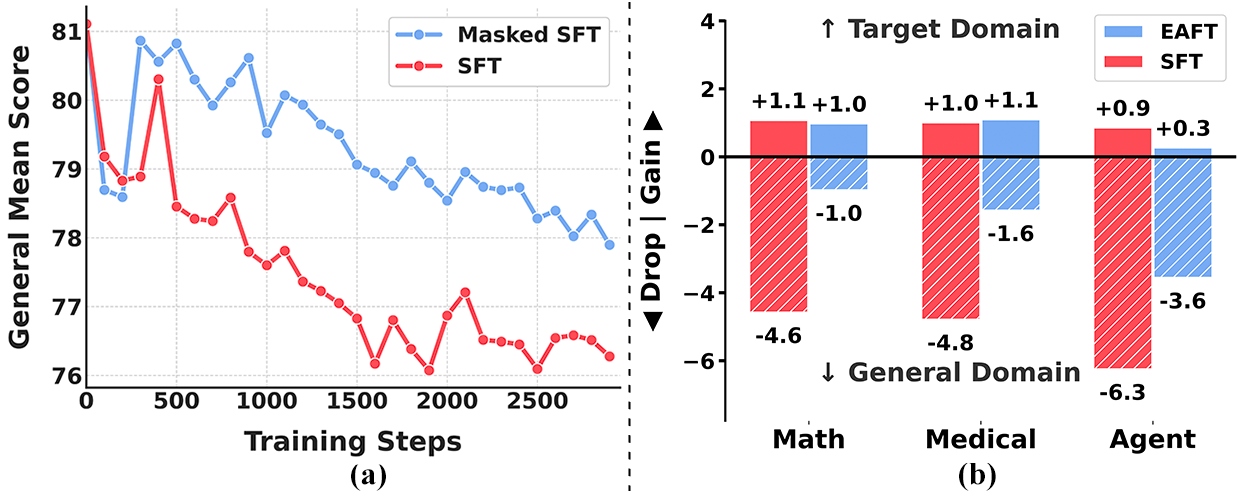}
    \vspace{-1em} 
    \caption{\textbf{(a)} Masking ``Confident Conflict'' tokens (the bottom 15\% in both entropy and probability) effectively mitigates the general capability degradation observed in standard SFT.
\textbf{(b)} Across Math, Medical, and Agent domains, EAFT matches SFT in target task improvements (upper bars) while significantly minimizing performance drops on general benchmarks (lower bars).}
\label{fig:main_2}

\end{figure}

To verify whether these conflicts are indeed the drivers of forgetting, we conducted a \textbf{pilot experiment}. By simply masking out these "Confident Conflict" tokens during training (Fig.~\ref{fig:main_2}). We observed that catastrophic forgetting was significantly mitigated compared to standard SFT.
This confirms that enforcing updates on these conflicting samples is the primary driver of capability degradation.

Building on this insight, we propose \textbf{Entropy-Adaptive Fine-Tuning (EAFT)}. Instead of using discrete thresholds, EAFT employs a soft gating mechanism that dynamically modulates the training loss based on token-level entropy.

Crucially, this approach differentiates EAFT from standard Cross-Entropy or probability-based re-weighting strategies~\citep{dft,talr,flow,rlRazor}. These methods rely solely on prediction probability, and thus risk amplifying destructive gradients on "Confident Conflicts."
In contrast, EAFT leverages entropy to distinguish rigidity from uncertainty. By down-weighting low-entropy tokens to suppress conflicting gradients, while concentrating supervision on high-entropy ones to facilitate adaptation, EAFT effectively balances domain proficiency with the preservation of general capabilities.

To verify the effectiveness and universality of our approach, we validate EAFT through extensive experiments on mathematical, extending to medical and agent domains. Our comprehensive evaluation covers diverse model families (Qwen, GLM) and scales ranging from 4B to 32B parameters. The results are presented in Fig. \ref{fig:main_2} and Tab.~\ref{tab:main_table}.

Quantitative results (Sec.~\ref{sec:main_result}) demonstrate that EAFT consistently outperforms standard SFT and existing mitigation strategies. It achieves a Pareto improvement: matching or exceeding baselines on target tasks while significantly mitigating catastrophic forgetting on general benchmarks.

Beyond performance, we provide an in-depth analysis of the method's intrinsic properties. We empirically verify that the entropy-adaptive mechanism successfully targets "Confident Conflicts" (Sec.~\ref{sec:mechanism}), and further demonstrate that EAFT is both robust to hyperparameter variations (Sec.~\ref{subsec:sensitivity}) and computationally efficient (Sec.~\ref{subsec:k_efficiency}).


In summary, our contributions are as follows:
\begin{itemize}[leftmargin=4mm]
\item We \textbf{uncover} the distinct distributional gap between SFT and on-policy RL data. Through visualization and pilot experiments, we pinpoint "Confident Conflicts" (low-entropy, low-probability tokens) as the primary cause of catastrophic forgetting.
\item We \textbf{propose} Entropy-Adaptive Fine-Tuning (EAFT), a novel objective that utilizes token-level entropy to modulate the training loss. This mechanism automatically down-weights destructive updates from conflicting data.
\item We \textbf{validate} our approach through extensive experiments on Math, Agent, and Medical domains. The results establish EAFT as an effective and universal solution that successfully mitigates catastrophic forgetting across diverse model families and scales (4B--32B).

\end{itemize}

\section{Related Work}\label{sec:related}

\textbf{Post-training Paradigms: SFT vs. RL.}
Post-training methods, primarily Supervised Fine-Tuning (SFT) and Reinforcement Learning (RL), are widely used to align pre-trained LMs \citep{qwen2.5,glm2024chatglm,yang2025qwen3}. SFT optimizes the model to maximize the likelihood of ground-truth demonstrations (Off-policy). By contrast, RL optimizes the model based on its own generated responses guided by reward signals (On-policy). These signals typically originate from parameterized reward models or verifiable signals~\citep{ppo,ouyang2022training,shao2024deepseekmath,zeng2025simplerl}.

Emerging research highlights a fundamental dichotomy in their learning behaviors. While SFT is efficient, it is inherently prone to memorization, often fits specific training samples at the expense of generalization \citep{chu2025sftmemorizes}. RL demonstrates superior robustness: it can benefit from single training examples without severe overfitting \citep{wang2025implicit}, and it updates a smaller, more effective subspace of parameters compared to SFT \citep{mukherjee2025reinforcement}. A common thread connecting these results is that RL parameter updates are more local and targeted \citep{razin2023vanishing}.

Our work investigates the root cause of SFT's instability. We argue that unlike on-policy methods which naturally operate within the model's distribution, standard SFT indiscriminately forces the fitting of "confident conflicts"—low-entropy samples that contradict the model's pre-trained knowledge.

\textbf{Catastrophic forgetting.}
Catastrophic forgetting remains a foundational challenge in neural networks \citep{mccloskey1989catastrophic,kirkpatrick2017overcoming}. Initial efforts to mitigate forgetting focused on preventing parameters from drastically changing \citep{kirkpatrick2017overcoming,li2017learning,lopez2017gradient}.

In LLM post-training, this manifests as the "Alignment Tax" \citep{askell2021general,ouyang2022training}: fine-tuning for domain-specific capabilities (e.g., mathematical problem solving, tool utilization, or biomedical adaptation) often significantly degrades the model's general capabilities \citep{ouyang2022training,luo2023empirical,shi2025continual}.

To overcome these limitations, recent works have explored dynamic training strategies that adjust optimization based on token-level metrics. TALR~\citep{talr} dynamically scales learning rates based on token confidence to accelerate convergence. DFT~\citep{dft} re-weights the SFT loss according to prediction probability. Others like RL's Razor \citep{rlRazor} employ KL divergence as a regularization term to constrain the model's drift from its base distribution.

However, existing dynamic methods predominantly rely on probability or KL divergence as proxies for difficulty or drift. We argue that probability alone is an insufficient statistic: a low-probability token can represent either epistemic uncertainty (valid knowledge to be learned) or a "Confident Conflict" (a destructive sample that contradicts the model's strong priors). By forcing the model to fit these conflicts based on probability, prior methods risk accelerating forgetting.
Our work advances this by introducing Entropy as a gating signal.

\section{Empirical Analysis \& Methodology}\label{sec:method}

In this section, we systematically investigate the causes of catastrophic forgetting in SFT and propose a targeted solution. We begin by defining the problem setup and key metrics in Sec.~\ref{subsec:preliminaries}. We then present an empirical analysis identifying `Confident Conflicts' as the primary source of destructive gradients in Sec.~\ref{subsec:analysis}. Finally, based on these insights, we introduce our method, \textbf{Entropy-Adaptive Fine-Tuning (EAFT)}, in Sec.~\ref{subsec:EAFT}.

\subsection{Preliminaries}
\label{subsec:preliminaries}

SFT is the standard process of adapting a base model $\theta$, denoted by its probability distribution $P_{\theta}$, to a target dataset $\mathcal{D} = \{(\boldsymbol{x}, \boldsymbol{y})_i\}_{i=1}^N$. For each sample, the response is a sequence of tokens $\boldsymbol{y} = (y_1, \dots, y_T)$, where $T$ denotes the sequence length. The adaptation is typically achieved by minimizing the Cross-Entropy (CE) loss, which maximizes the likelihood of the target sequences:

\begin{equation}
\mathcal{L}_{\text{CE}}(\theta) = - \sum_{t=1}^{T} \log P_{\theta}(y_t | \boldsymbol{x}, \boldsymbol{y}_{<t})
\label{eq:ce_loss}
\end{equation}

A key limitation of this objective is its uniform treatment of all tokens. It aggressively updates model parameters to fit every token $y_t$ regardless of the model's prior knowledge or uncertainty.

To investigate the dynamics of how this uniform objective interacts with the model's internal state, we introduce two token-level metrics that serve as the foundation for our analysis and method:
\begin{enumerate}[leftmargin=4mm]
    \item \textbf{Probability.} $p_t = P_{\theta}(y_t | \boldsymbol{x}, \boldsymbol{y}_{<t})$, represents model's confidence in the ground-truth token.
    
    \item \textbf{Predictive Entropy.} Let $P_t(v) \triangleq P_{\theta}(v | \boldsymbol{x}, \boldsymbol{y}_{<t})$ denote the distribution at step $t$. The entropy is defined as:
        $H_t = -\sum_{v \in \mathcal{V}} P_t(v) \log P_t(v)$
    This measures the model's predictive uncertainty over the vocabulary $\mathcal{V}$.
\end{enumerate}

\subsection{Analysis: The Origins of Forgetting}
\label{subsec:analysis}

To understand why SFT leads to forgetting while on-policy RL does not, we compare the token-level statistics of standard SFT data against model-generated rollouts (the data source for on-policy RL). We visualize the distribution of probability $p_t$ and entropy $H_t$ for both datasets in Fig.~\ref{fig:main}.

\paragraph{Distributional Gap: Confident Conflicts.}
The visualization reveals a critical distributional shift. On-policy data falls into either high-probability (model is correct) or high-entropy (model is exploring).
In sharp contrast, SFT data contains a significant cluster of tokens with both \textbf{Low Entropy ($H_t \downarrow$) and Low Probability ($p_t \downarrow$)}.
We term these samples \textbf{`Confident Conflicts'}. They represent cases where the model holds a strong, stubborn prior belief (low entropy) that directly contradicts the ground-truth label (low probability).

\paragraph{Pilot Study: Masking Confident Conflicts.}

We hypothesize that these `Confident Conflicts' are the primary drivers of forgetting. To verify this, we conducted a pilot experiment where we masked the loss for tokens falling within the \textbf{bottom 15\% of both entropy and probability rankings}. As shown in Fig.~\ref{fig:main_2}, this simple intervention significantly mitigates the general capability degradation observed in standard SFT.

Notably, masking these specific tokens nearly eliminated catastrophic forgetting on our benchmarks. This finding confirms that the degradation of general capabilities stems primarily from forcing the model to accommodate these conflicting samples, rather than from the SFT process itself.

\paragraph{Theoretical Insight.}
We analyze the optimization dynamics to understand this damage. Consider the CE loss (Eq.~\ref{eq:ce_loss}).
When the model is highly confident in a prediction that contradicts the target (low entropy, low probability), the CE loss induces a very large gradient. Because the model strongly favors another token, fitting the target requires substantial parameter updates, which can overwrite general representations in the base model. By contrast, when the model is uncertain (high entropy), the gradients are smaller and updates are gentler, helping preserve the model’s original capabilities.

\subsection{Entropy-Adaptive Fine-Tuning (EAFT)}
\label{subsec:EAFT}

While the pilot study validates our hypothesis, the hard masking strategy has two limitations: it discards training data, leads to ineffective learning on the target domain, and it relies on sensitive hyperparameters ($\tau, \delta$).
To address this, we propose \textbf{Entropy-Adaptive Fine-Tuning (EAFT)}, a soft gating mechanism that dynamically adjusts the learning signal based on the model's uncertainty.

\paragraph{The EAFT Objective.}
We formulate the EAFT loss by scaling the standard supervision with the normalized entropy. This mechanism \textbf{prioritizes learning} from samples where the model is exploring, while effectively suppressing the gradients when the model is confident but conflicting. The objective is decomposed as:
\vspace{-1mm}
\begin{equation}
\mathcal{L}_{\text{EAFT}}(\theta) = - \sum_{t=1}^{T} \underbrace{\tilde{H}_t}_{\substack{\text{Adaptive} \\ \text{Gating Signal}}} \cdot \underbrace{\log P_{\theta}(y_t | \boldsymbol{x}, \boldsymbol{y}_{<t})}_{\text{Standard Supervision}}
\label{eq:EAFT_loss}
\end{equation}

Here, the gating term $\tilde{H}_t$ is derived from the entropy of the \textbf{Top-$K$ tokens}. This approximation greatly reduces computation compared with using the full vocabulary (analysis in Sec.~\ref{subsec:k_efficiency}). Normalized to the range $[0, 1]$ with $K=20$, we calculate:
\vspace{-1mm}
\begin{equation}
\tilde{H}_t = \frac{H_t^{\text{top-}K}}{\ln(K)} \approx \frac{H_t^{\text{top-}20}}{3.0}
\end{equation}

where $H_t^{\text{top-}K}$ denotes the entropy calculated over the top-$K$ probability distribution, and $\ln(K)$ serves as the normalization factor (the maximum entropy for $K$ outcomes).
This normalization creates a self-regulating mechanism:
\begin{itemize}[leftmargin=4mm]
    \vspace{-0.2em}
    \item \textbf{Conflict Suppression ($\tilde{H}_t \to 0$):} When the model is stubborn (low entropy), the weight drops, effectively \textbf{masking} the destructive gradient from the conflicting label.
    \vspace{-0.5em}
    \item \textbf{Knowledge Acquisition ($\tilde{H}_t \to 1$):} When the model is uncertain (high entropy) or exploring, the weight remains high, recovering the standard SFT objective to learn new patterns.
    \vspace{-1em}
\end{itemize}

\begin{table*}[t!]
\centering
\vspace{-0.5em}
\setlength\tabcolsep{1.5pt}
\renewcommand{\arraystretch}{1}
\fontsize{8.1pt}{10.5pt}\selectfont
\begin{tabular}{p{2.8cm}cccccccc}
\toprule
\multirow{2}[2]{*}{\textbf{Method}} & \multicolumn{3}{c}{\textbf{Math Domain}} & \multirow{2}[2]{*}{\textbf{Math Avg.}} & \multicolumn{3}{c}{\textbf{General Domain}} & \multirow{2}[2]{*}{\textbf{General Avg.}} \\
\cmidrule(lr){2-4} \cmidrule(lr){6-8}
& AIME24 & AIME25 & GSM8K &  & MMLU & IFEval & CLUEWSC & \\
\midrule
\textbf{Qwen3-4B-Instruct} & 63.3 & 47.4 & 94.3 & 68.3 & 77.1 & 81.0 & 85.2 & 81.1 \\
\quad + SFT & \underline{63.3} & \underline{50.0} & \textbf{94.8} & \textbf{69.4} & \underline{76.5} & \underline{79.5} & 74.5 & 76.5 {\color{red}(-4.6)}\\
\quad + $\mathrm{SFT}_{\mathrm{KL}}$ & \underline{63.3} & \underline{50.0} & 93.6 & 69.0 & 74.5 & 74.9 & \textbf{89.4} & \underline{79.6} {\color{red}(-1.5)}\\
\quad + FLOW & \textbf{66.7} & 46.7 & 94.3 & 69.2 & 76.2 & 78.3 & 82.8 & 79.1 {\color{red}(-2.0)}\\
\quad + DFT & 56.7 & 40.0 & 93.9 & 63.5 & 75.9 & 77.0 & 81.4 & 78.1 {\color{red}(-3.0)}\\
\quad + TALR & 50.0 & \underline{50.0} & 93.3 & 64.4 & 76.2 & 78.1 & 74.5 & 76.2 {\color{red}(-4.9)}\\
\rowcolor[RGB]{236,244,252} \quad + EAFT & 60.0 & \textbf{53.3} & \underline{94.5} & \underline{69.3} & \textbf{76.6} & \textbf{80.1} & \underline{83.7} & \textbf{80.1} {\color{red}(-1.0)}\\
\midrule
\textbf{Qwen2.5-32B-Instruct} & 22.2 & 13.3 & 96.0 & 43.8 & 84.1 & 78.3 & 91.9 & 84.8 \\
\quad + SFT & \textbf{53.3} & \textbf{50.0} & \underline{96.3} & \textbf{66.5} & 76.9 & 74.2 & 93.8 & 81.6 {\color{red}(-3.2)}\\
\quad + $\mathrm{SFT}_{\mathrm{KL}}$ & 33.3 & 33.3 & 94.1 & 53.6 & \textbf{81.4} & 68.1 & 93.2 & 80.9 {\color{red}(-3.9)}\\
\quad + FLOW & \underline{50.0} & \textbf{50.0} & \underline{96.3} & 65.4 & 78.6 & \underline{75.1} & 93.6 & \underline{82.4} {\color{red}(-2.4)}\\
\quad + DFT & 33.3 & 36.7 & 95.9 & 55.3 & 77.8 & 70.0 & \textbf{94.4} & 80.7 {\color{red}(-4.1)}\\
\quad + TALR & 40.0 & 43.3 & 95.3 & 59.5 & 73.1 & 72.5 & \underline{94.1} & 79.9 {\color{red}(-4.9)}\\
\rowcolor[RGB]{236,244,252} \quad + EAFT & \textbf{53.3} & \underline{46.7} & \textbf{96.5} & \underline{65.5} & \underline{79.0} & \textbf{78.4} & 93.9 & \textbf{83.7} {\color{red}(-1.1)}\\
\midrule
\textbf{GLM4-9B-0414} & 6.7 & 6.7 & 90.1 & 34.5 & 70.2 & 74.4 & 85.1 & 76.6 \\
\quad + SFT & \textbf{20.0} & \underline{10.0} & 90.3 & \underline{40.1} & 57.3 & 69.8 & 84.8 & 70.6 {\color{red}(-6.0)}\\
\quad + $\mathrm{SFT}_{\mathrm{KL}}$ & 13.3 & 6.7 & 90.1 & 36.7 & \underline{60.0} & 66.4 & \underline{85.3} & 70.5 {\color{red}(-6.1)}\\
\quad + FLOW & \underline{16.7} & \textbf{13.3} & 91.1 & \textbf{40.4} & 57.5 & \underline{71.5} & 85.2 & 71.4 {\color{red}(-5.2)}\\
\quad + DFT & 13.3 & 6.7 & 89.0 & 36.4 & 48.9 & 69.7 & \textbf{86.0} & 68.2 {\color{red}(-8.4)}\\
\quad + TALR & 15.6 & \textbf{13.3} & \underline{91.2} & 40.0 & 57.4 & 71.3 & 84.5 & \underline{71.5} {\color{red}(-5.1)}\\
\rowcolor[RGB]{236,244,252} \quad + EAFT & 13.3 & \textbf{13.3} & \textbf{91.5} & 39.4 & \textbf{60.8} & \textbf{72.0} & \underline{85.3} & \textbf{72.7} {\color{red}(-3.9)} \\
\bottomrule
\vspace{-4em}
\end{tabular}
\vspace{-2em}
\caption{Main results on the target domain (Math) and general domain benchmarks. We evaluate performance on AIME24, AIME25 and GSM8K as the training target, alongside MMLU, IFEval, and CLUEWSC for general capabilities. The top two outcomes are \textbf{bolded} and \underline{underlined}. All results are averaged over three independent runs. The ``Avg.'' represents the average performance of the datasets in the corresponding domain.}
\label{tab:main_table}
\end{table*}

\section{Experiments}

In this section, we empirically validate the effectiveness of EAFT. Our experiments are designed to answer the following three key research questions:
\begin{itemize}[leftmargin=4mm]
    \item RQ1 (Performance): Can EAFT mitigate catastrophic forgetting without compromising performance on the target task?
    \vspace{-0.2cm}
    \item RQ2 (Mechanism): Does the entropy-adaptive gating mechanism work as intended in filtering “Confident Conflict” samples?
    \vspace{-0.2cm}
    \item RQ3 (Generalization): Is the efficacy of EAFT inherently domain-agnostic?
\end{itemize}

\subsection{Experimental Settings}
\textbf{Datasets.} We utilize prompts from NuminaMath~\citep{li2024numinamath}, BigMathVerified~\citep{albalak2025big}, and Nemotron-CrossThink~\citep{akter2025nemotron}, synthesizing responses via Qwen3-235B-A22B-Instruct \citep{yang2025qwen3}. We randomly select 19k correctly answered data pairs as our math training data. Details in App.~\ref{app:datasets} \\
\textbf{Benchmarks.} Our evaluation covers the target domain (Math: AIME 24/25~\citep{aime}, GSM8K~\citep{GSM8K}) and general domain (MMLU~\citep{mmlu}, IFEval~\citep{ifeval}, CLUEWSC~\citep{xu2020clue}). We report the average score (``Avg'') across benchmarks, with full details available in App.~\ref{app:benchmarks}.\\
\textbf{Models.} To verify the scalability and generalizability of our method, we conduct experiments across a diverse model zoo spanning multiple families and parameter scales. Specifically, our evaluation includes models ranging from 4B to 32B parameters: Qwen3-4B-Instruct \citep{yang2025qwen3}, GLM4-9B-0414 \citep{glm2024chatglm}, and Qwen2.5-32B-Instruct \citep{qwen2.5}. Details in App.~\ref{app:models} \\
\textbf{Baselines.} We compare our proposed method against standard Supervised Fine-Tuning (SFT) and a regularized variant, denoted as $\text{SFT}_{\text{KL}}$, which incorporates a Kullback-Leibler (KL) divergence constraint into the loss function to prevent model drift. Additionally, we include several advanced alignment baselines, including FLOW \citep{flow}, DFT \citep{dft}, and TALR \citep{talr}. Details and implementation settings are provided in App.~\ref{app:baselines} and App.~\ref{app:implementation}

\subsection{Main Results (Answering RQ1)}
\label{sec:main_result}
Tab.~\ref{tab:main_table} presents the performance comparison across diverse model families and scales. The results provide a compelling answer to RQ1: \textbf{EAFT maintains competitive performance on target domain (Math) while significantly mitigating catastrophic forgetting on general capabilities.}

\textbf{General Capabilities.}
The main advantage of EAFT lies in stability. Standard SFT causes a sharp drop in general benchmarks (e.g., a 10.7 point drop on CLUEWSC for Qwen3-4B). In contrast, EAFT effectively preserves the model's original knowledge, achieving the highest average score across general tasks among all baselines. This demonstrates that EAFT offers the most robust capability retention compared to other methods.

\textbf{Target Domain Performance.}
EAFT achieves highly competitive results on the target domain compared to other baselines. Specifically, the gap between EAFT and the best-performing method on math score is consistently less than 1 point. Notably, EAFT achieves the best performance on several sub-benchmarks. This demonstrates that EAFT effectively adapts to target domain, maintaining the same level of learning capability as standard SFT.

\begin{figure}[t]
    \centering\includegraphics[width=0.99\linewidth]{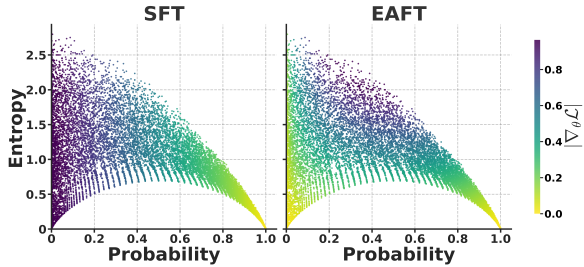}
    \caption{Gradient Magnitude Landscape.
\textbf{Left:} SFT exerts strong optimization pressure (dark purple) on \textit{Confident Conflicts} in the bottom-left. \textbf{Right:} EAFT effectively suppresses these gradients (light yellow), protecting the model's existing representations.}
    \label{fig:scatter}
    \vspace{-0.8em}
\end{figure}

\subsection{Mechanism Analysis (Answering RQ2)}
\label{sec:mechanism}
To understand the inner workings of EAFT, we analyze the training dynamics from both spatial and temporal perspectives. The results provide a clear affirmative answer to RQ2: \textbf{EAFT effectively filters out "Confident Conflict" samples, preventing destructive gradient updates.}

\textbf{Gradient Landscape.}
Fig.~\ref{fig:scatter} visualizes the optimization strength for each token, where color intensity represents the gradient magnitude.
The gradient distribution exhibits a distinct skew rather than uniformity. Due to the nature of Cross-Entropy loss, low-probability tokens are subjected to the strongest optimization pressure (indicated by the dark density in the left region).
Crucially, in the "Confident Conflict" zone (bottom-left), the model is forcefully updated to fit labels that contradict its confident priors. These aggressive updates destabilize the model's established representation space.
In contrast (right), EAFT dramatically alters this landscape. The "Confident Conflict" region becomes pale, indicating near-zero gradients.
Here, the low entropy term directly suppresses the high gradients generated by the Cross-Entropy loss. This protects the model's established knowledge from being over-optimized by conflicting data.

\begin{figure}[t]
    \centering\includegraphics[width=0.5\textwidth]{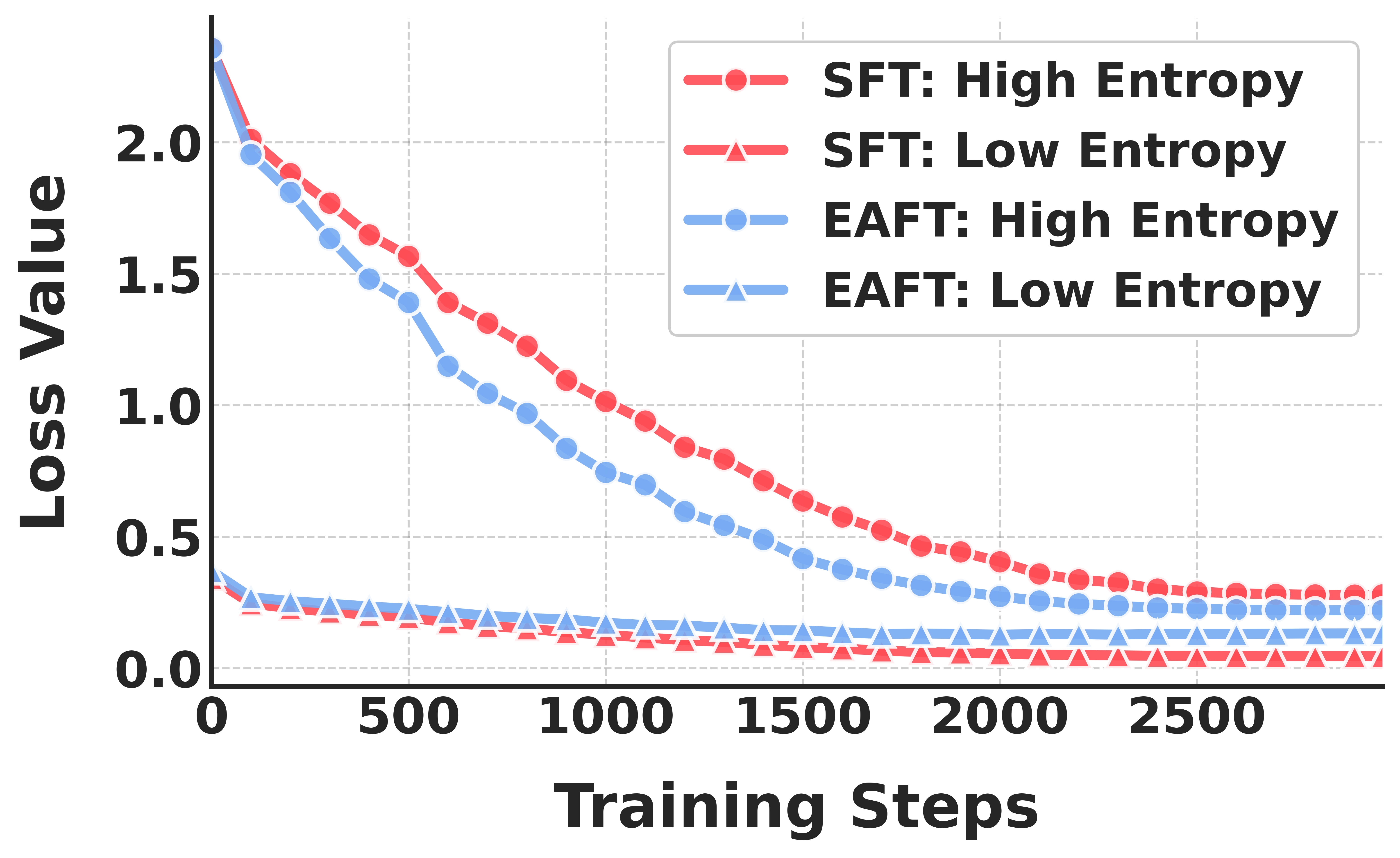}
    \caption{Training dynamics of token subgroups. EAFT matches SFT on high-entropy tokens while keeping losses stable on low-entropy conflicts, preventing over-optimization of conflicting priors. High and low entropy correspond to values $\ge 2.0$ and $\le 0.5$, respectively.}
    \label{fig:dynamics}
    \vspace{-0.8em}
\end{figure}

\textbf{Training Dynamics.}
Fig.~\ref{fig:dynamics} tracks the loss of cross-entropy of different token types throughout the training process, comparing EAFT (Blue) with Standard SFT (Red). We categorize tokens into high-entropy and low-entropy groups. \textbf{High-Entropy Tokens} (marked with dots $\bullet$): EAFT exhibits a rapid loss reduction comparable to SFT. This confirms that our entropy-based gating effectively optimizes high-entropy tokens, ensuring the model adapts to the target domain without hindrance.
\textbf{Low-Entropy Tokens} (marked with triangles $\blacktriangle$): SFT (Red) aggressively drives this loss toward zero, indicating that the model is being forced to memorize data that conflicts with its priors.
In contrast, EAFT (Blue) maintains a stable loss throughout training. This demonstrates that the mechanism successfully prevents over-optimization on "Confident Conflict" tokens.

\begin{table*}[t]
\centering
\vspace{-1.5em}
\setlength\tabcolsep{2.5pt}
\renewcommand{\arraystretch}{1.1}
\fontsize{8.5pt}{10.5pt}\selectfont
\begin{tabular}{p{3.2cm}cccccccc}
\toprule
\multirow{2}[2]{*}{\textbf{Method}} & \multicolumn{3}{c}{\textbf{Medical Domain}} & \multirow{2}[2]{*}{\textbf{Medical Avg.}} & \multicolumn{3}{c}{\textbf{General Domain}} & \multirow{2}[2]{*}{\textbf{General Avg.}} \\
\cmidrule(lr){2-4} \cmidrule(lr){6-8}
 & MedMCQA & MedQA & PubMedQA &  & MMLU & IFEval & CLUEWSC & \\
\midrule
Qwen3-4B-Thinking & \underline{63.5} & 78.2 & 76.0 & 72.6 & \underline{79.3} & \textbf{85.0} & \textbf{94.1} & \textbf{86.1} \\
\quad + SFT & 63.3 & \underline{79.5} & \textbf{78.0} & \underline{73.6} & 78.3 & 75.3 & 90.4 & 81.3 {\color{red}(-4.8)} \\
\rowcolor[RGB]{236,244,252} \quad + EAFT & \textbf{63.9} & \textbf{80.0} & \underline{77.2} & \textbf{73.7} & \textbf{80.1} & \underline{81.7} & \underline{91.8} & \underline{84.5} {\color{red}(-1.6)} \\
\bottomrule
\vspace{-2em}
\end{tabular}
\vspace{-1.5em}
\caption{Results on target (Medical) and General domain benchmarks. We evaluate performance on MedMCQA, MedQA, and PubMedQA for the medical domain, alongside MMLU, IFEval, and CLUEWSC for general capabilities. The top two outcomes are \textbf{bolded} and \underline{underlined}. All results are averaged over three independent runs.}
\label{tab:medical_general_results}
\end{table*}

\begin{table*}[t]
\centering
\setlength\tabcolsep{4pt}
\renewcommand{\arraystretch}{1.1}
\fontsize{8.5pt}{10.5pt}\selectfont
\begin{tabular}{p{3.5cm}ccccc}
\toprule
\multirow{2}[2]{*}{\textbf{Method}} & \textbf{Agent Domain} & \multicolumn{3}{c}{\textbf{General Domain}} & \multirow{2}[2]{*}{\textbf{General Avg.}} \\
\cmidrule(lr){2-2} \cmidrule(lr){3-5}
 & BFCL v3 & MMLU & IFEval & CLUEWSC & \\
\midrule
Qwen3-4B-Instruct & 60.5 & \textbf{77.1} & \textbf{81.0} & \textbf{85.2} & \textbf{81.1} \\
\quad + SFT & \textbf{61.4} & 74.5 & 77.8 & 72.2 & 74.8 {\color{red}(-6.3)} \\
\rowcolor[RGB]{236,244,252} \quad + EAFT & \underline{60.8} & \underline{76.1} & \underline{78.6} & \underline{77.7} & \underline{77.5} {\color{red}(-3.6)}\\
\bottomrule
\vspace{-2em}
\end{tabular}
\vspace{-1em}
\caption{Results on target (Agent Toolcall) and General domain benchmarks. We evaluate performance on BFCL for the agent toolcall domain, alongside MMLU, IFEval, and CLUEWSC for general capabilities. The top two outcomes are \textbf{bolded} and \underline{underlined}. All results are averaged over three independent runs.}
\label{tab:agent_general_results}
\vspace{-1em}
\end{table*}

\subsection{Universality (Answering RQ3)}

To verify whether the efficacy of EAFT is domain-agnostic, we extend our evaluation to two distinct specialized domains: Biomedical (knowledge-intensive) and Agent Tool-Use (syntax-intensive). The results clearly answer RQ3: \textbf{EAFT is a domain-agnostic solution that consistently mitigates catastrophic forgetting across tasks.}

\textbf{Medical Domain.} We first examine the medical domain, a field demanding domain-specific knowledge application. 
For this experiment, we fine-tune the \textbf{Qwen3-4B-Thinking} \citep{yang2025qwen3} model using the \textbf{Huatuo-O1} \citep{2412_HuatuoGPT_o1} dataset and conduct evaluations on the \textbf{MedMCQA} \citep{pal2022medmcqa},\textbf{PubMedQA} \citep{jin2019pubmedqa}and \textbf{MedQA} \citep{medqa} benchmarks (see App. \ref{app:benchmarks} for more details).
In Tab. \ref{tab:medical_general_results}, standard SFT triggers severe catastrophic forgetting while adapting to the target domain, causing average performance on general benchmarks to drop significantly from 86.1 to 81.3. EAFT effectively mitigates this issue. It not only preserves general capabilities (maintaining an average of 84.5) but also marginally outperforms standard SFT on the target medical tasks (73.7 vs. 73.6). This result suggests that EAFT can inject specific knowledge without destructively overwriting the model's core representation.

\textbf{Agent Tool-Use.} We further evaluate EAFT on agentic tool-use tasks, which require strict adherence to syntactic constraints and formats. 
Specifically, we utilize the subset of \textbf{Nemotron-Agentic-Tool-Use-v1}~\citep{nvidia_nemotron_agentic} to train the \textbf{Qwen3-4B-Instruct} ~\citep{yang2025qwen3} and assess its performance on the Berkeley Function Calling Leaderboard (\textbf{BFCL})~\citep{BFCL} (see App.~\ref{app:benchmarks}).
The results in Tab. \ref{tab:agent_general_results} reveal a similar pattern regarding robustness. Although SFT fits the target distribution aggressively to achieve a slightly higher score (61.4), it does so at the cost of catastrophic forgetting, resulting in a sharp decline in general capabilities (81.1 $\rightarrow$ 74.8). In contrast, EAFT achieves a far superior balance: it remains highly competitive on the target task (60.8, within 1\% of SFT) while maintaining general performance (77.5). These experiments confirm that EAFT is a domain-agnostic solution that consistently alleviates catastrophic forgetting across varying data distributions.

\section{Analysis and Discussion}
\label{sec:analysis}
Having demonstrated the superior performance of EAFT, this section investigates the \textbf{intrinsic robustness} and \textbf{computational efficiency} of the proposed mechanism. Specifically, we conduct ablation studies to answer two fundamental questions:

\vspace{-2mm}
\begin{itemize}[leftmargin=4mm]
    \item \textbf{Robustness:} Does performance rely on linear entropy scaling, or is the "entropy-aware" mechanism the primary driver? (Sec.~\ref{subsec:sensitivity})
    \item \textbf{Efficiency:} Can we approximate the token-level entropy efficiently without incurring significant computational overhead? (Sec.~\ref{subsec:k_efficiency})
\end{itemize}

\subsection{Robustness to Gating Function Variations}
\label{subsec:sensitivity}

In Sec.~\ref{sec:method}, we adopted a linear gating function $\mathcal{L} = \tilde{H}_t \cdot \mathcal{L}_{\text{CE}}$. To verify that our gains stem from the intrinsic \textbf{mechanism of entropy awareness} rather than a specific hyperparameter choice, we generalize the gating signal to $f(\tilde{H}_t)$ and evaluate three distinct categories of variants:

\begin{itemize}[leftmargin=4mm]
    \vspace{-0.5em}
    \item \textbf{Polynomial Scaling:} We test stricter suppression of confident samples using power functions $f(\tilde{H}_t) = (\tilde{H}_t)^p$ with $p \in \{2, 3\}$. We denote these variants as \textbf{EAFT$^2$} and \textbf{EAFT$^3$}.
    \vspace{-0.5em}
    \item \textbf{Non-linear Gating (Sigmoid):} We employ a Sigmoid activation $f(\tilde{H}_t) = \sigma\big(\alpha (\tilde{H}_t - \beta)\big)$ where $\alpha$ controls the steepness and $\beta$ determines the centering threshold. In our experiments, we set $\beta=0.17$ (aligning with the bottom 15\% entropy percentile used in the Masked SFT baseline) and $\alpha=30$. This variant is denoted as \textbf{EAFT$_{sig}$}. See App.~\ref{app:discussion} for further comparisons.
    \vspace{-0.5em}
    \item \textbf{Piecewise Hard Thresholding (Hard Mask):} We implement a binary filter that strictly discards the bottom 15\% lowest entropy tokens:
    \begin{equation}
        f(\tilde{H}_t) = \mathbb{I}(\tilde{H}_t > \tau_{0.15})
    \end{equation}
    where $\mathbb{I}(\cdot)$ is the indicator function and $\tau_{0.15}$ denotes the threshold value corresponding to the 15\textsuperscript{th} percentile of the entropy distribution. This sets the loss of "confident" tokens to zero, retaining only uncertain samples for training. We refer to this baseline as \textbf{Masked SFT}.
\end{itemize}
\vspace{-0.5em}

\begin{figure}[t]
    \centering
    \includegraphics[width=0.9\linewidth]{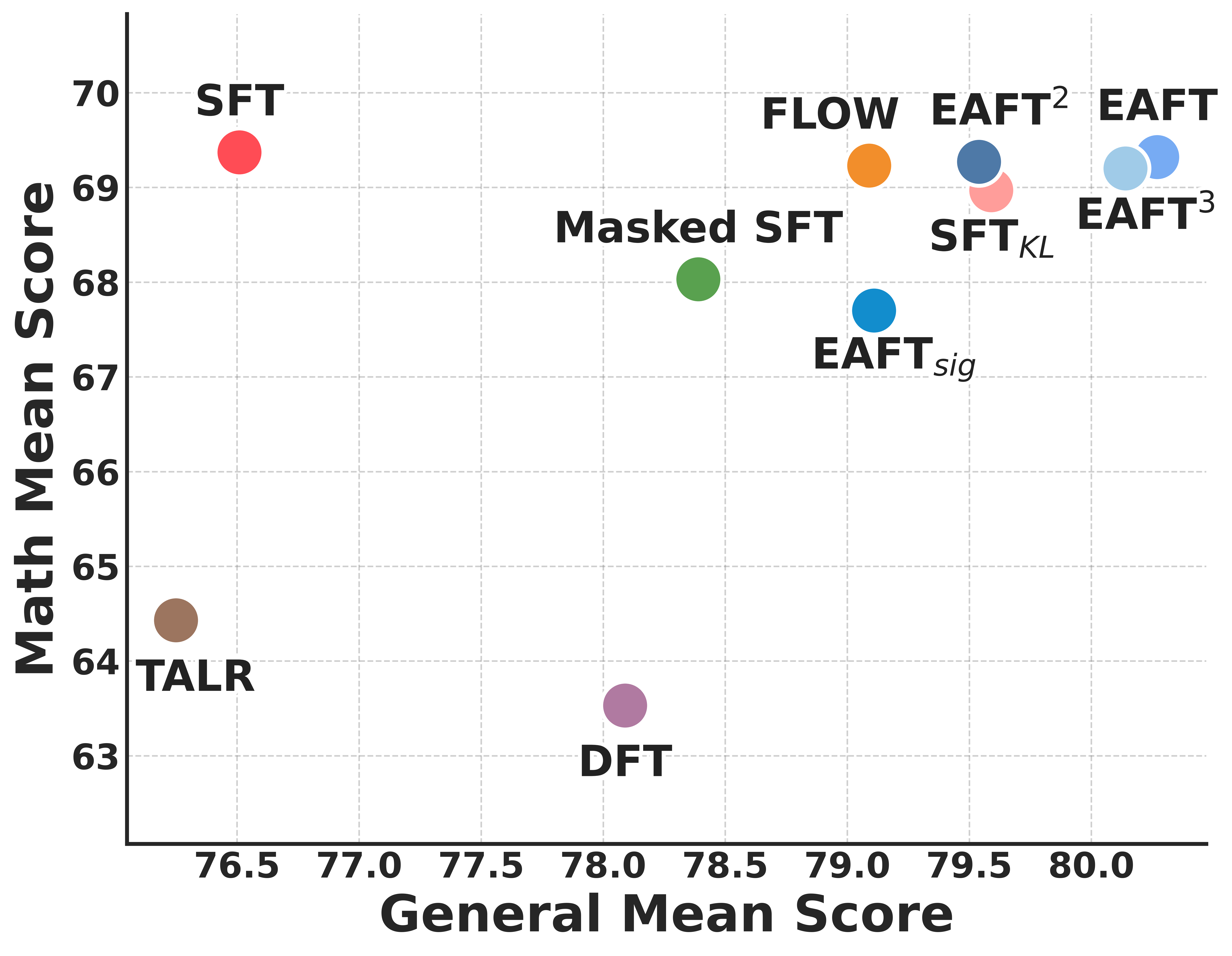}
    \caption{\textbf{Pareto trade-off analysis.} 
Unlike Masked SFT (drop in target score) or Standard SFT (severe forgetting), EAFT variants consistently occupy the optimal \textbf{top-right frontier}. This confirms that soft entropy-gating effectively preserves general capabilities without compromising target domain adaptation.}
    \label{fig:pareto_scatter}
\vspace{-0.8em}
\end{figure}

\textbf{Results and Discussion.} 
Fig.~\ref{fig:pareto_scatter} visualizes the trade-off between target domain performance (math average score) and general average score across all variants. We observe two pivotal insights:

\textbf{Universality of Entropy Awareness.} 
All entropy-aware variants (EAFT, EAFT$^2$, EAFT$^3$, EAFT$_{sig}$) consistently outperform SFT in general capabilities. This confirms that the reduction in forgetting stems from the core \textbf{mechanism of entropy monitoring} rather than the specific mathematical form. Simply down-weighting high-conflict tokens is sufficient to preserve general abilities.

\textbf{The Necessity of Soft Gating.} 
We observe a critical pitfall in the \textbf{Mask} baseline: while strictly discarding confident tokens prevents forgetting, it significantly harms target task performance (Math Score drops to 65.60 vs. EAFT's 69.27). This implies that "confident conflicts" carry essential adaptation signals that a hard cutoff destroys. In contrast, EAFT's soft gating reduces their impact without removing them, successfully occupying the \textbf{Pareto frontier} of learning versus retaining.

\begin{figure}[t]
    \centering
    \includegraphics[width=0.98\linewidth]{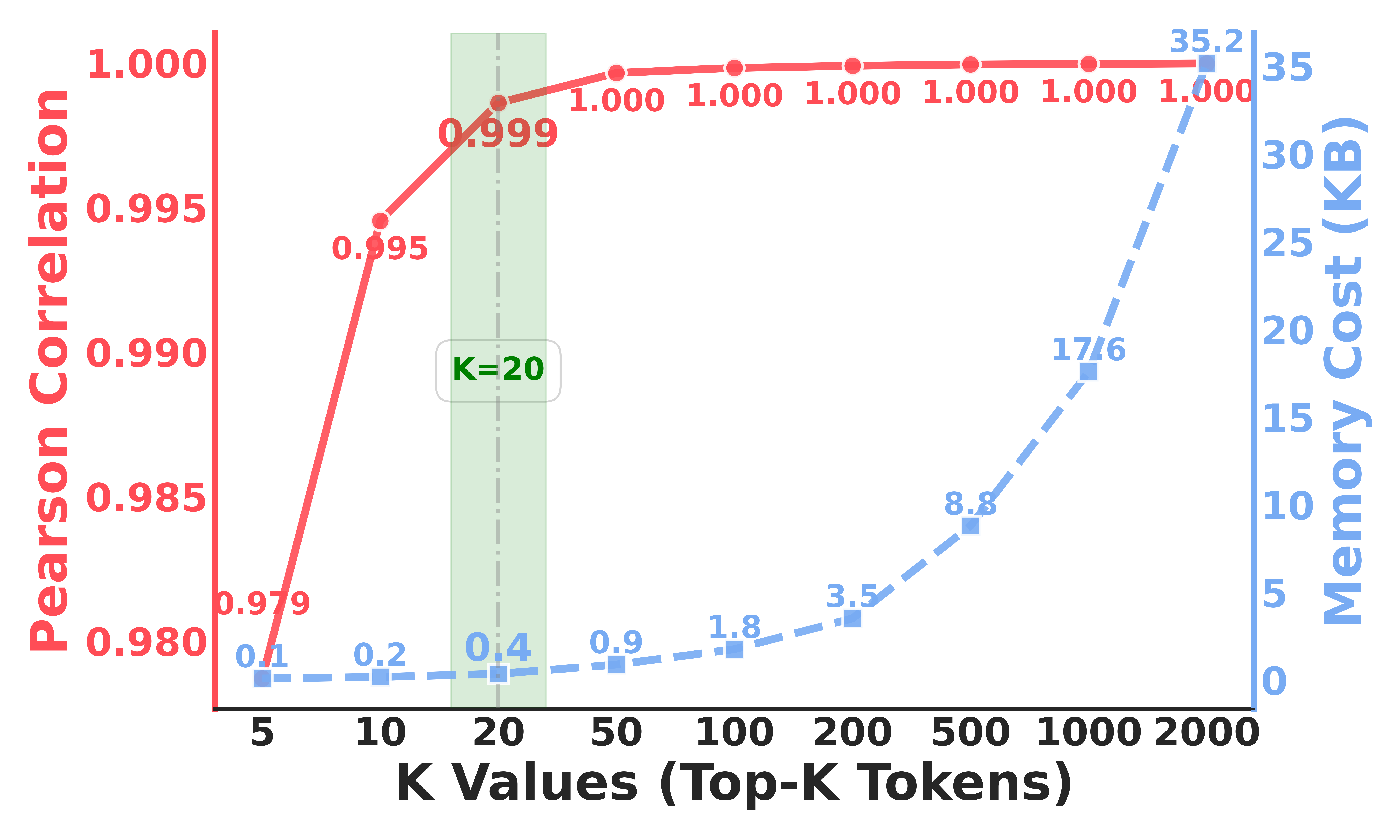}
    \caption{\textbf{Trade-off between Approximation Accuracy and Memory Cost.} The Red solid line (Left Axis) shows the Pearson correlation between Top-$K$ and exact entropy, which saturates rapidly at $0.999$. The Blue dashed line (Right Axis) tracks the additional memory overhead. The green shaded region highlights $K=20$ as the optimal operating point, achieving near-perfect fidelity with negligible computational cost.}
    \label{fig:k_tradeoff}
    \vspace{-0.8em}
\end{figure}

\subsection{Efficiency of Top-$K$ Approximation}
\label{subsec:k_efficiency}
Computing exact entropy over the entire vocabulary (often $|\mathcal{V}| > 100k$) creates unnecessary computational bottlenecks. In Sec.~\ref{sec:method}, we proposed a Top-$K$ approximation strategy ($\tilde{H} \approx H^{\text{top-}K}$), premised on the property that the probability mass of LLMs is highly sparse and concentrated in the leading tokens~\citep{fan2018hierarchical,radford2019language,holtzman2019curious}.

Fig.~\ref{fig:k_tradeoff} quantitatively validates this design by plotting approximation fidelity (Left Axis, Red) against memory cost (Right Axis, Blue). We observe a decisive trade-off: as $K$ increases, the Pearson correlation with the exact entropy rises sharply and rapidly plateaus. Specifically, at $K=20$ (green zone), the correlation reaches \textbf{0.999}, confirming that the long-tail distribution contributes negligibly to entropy. Conversely, the memory cost for the required operations (sort and log-sum-exp) remains virtually zero ($< \mathbf{0.4}$ KB) for $K \le 20$. Based on these findings, \textbf{we adopt the Top-20 approximation to estimate entropy.} This setting provides an accurate estimate of the full-vocabulary entropy, while introducing no noticeable computational overhead compared to standard SFT.

\section{Conclusion}

In this work, we identify "Confident Conflicts" as the primary driver of catastrophic forgetting in SFT. We introduce Entropy-Adaptive Fine-Tuning (EAFT), a method that dynamically modulates training loss based on token-level entropy. By suppressing gradients from conflicting data, EAFT effectively prevents destructive updates while maintaining learning efficiency. Extensive experiments across diverse domains and model scales validate our approach. Ultimately, EAFT provides a simple yet robust solution for balancing domain adaptation with the preservation of general capabilities.

\section{Limitations}

\textbf{Scope of Applicability (Counterfactual Scenarios).} 
It is important to clarify that EAFT is designed primarily for \textit{domain adaptation} and \textit{continual learning}, where the goal is to extend the model's capabilities without erasing existing knowledge. It is not a universal replacement for standard SFT, particularly in scenarios requiring \textbf{knowledge editing} or \textbf{counterfactual training} (e.g., teaching the model that ``the sky is green'' or correcting outdated facts). In such cases, the model's resistance to "Confident Conflicts" is undesirable, as the objective is precisely to override the prior belief. EAFT would interpret these necessary updates as conflicts and suppress them, hindering the intended learning.

\textbf{Target Performance Trade-off.} 
While EAFT successfully mitigates catastrophic forgetting, it does not aim to surpass standard SFT on the target domain metrics. As observed in our experiments, EAFT achieves a Pareto improvement—maintaining high general capabilities while closely approaching, but not necessarily exceeding, the peak specialization performance of SFT. For applications where maximizing target domain performance is the sole priority, regardless of the degradation in other areas, standard SFT may still be the preferred choice.

\textbf{Dependence on Base Model Quality.} 
EAFT operates on the assumption that the model's high-confidence priors represent valuable general knowledge worth preserving. However, if the base model exhibits \textbf{miscalibrated confidence} (e.g., being confidently wrong or hallucinating), EAFT inadvertently protect these erroneous behaviors. Future work could explore incorporating uncertainty calibration techniques to distinguish between true knowledge and confident hallucinations.

\clearpage
\newpage
\appendix
\section*{Appendix}
\startcontents[sections]
\printcontents[sections]{l}{1}{\setcounter{tocdepth}{3}}

\section{Qualitative Analysis: What are "Confident Conflicts"?}
\label{app:entropy_analysis}

\begin{figure}[t]
    \centering
    \includegraphics[width=0.95\linewidth]{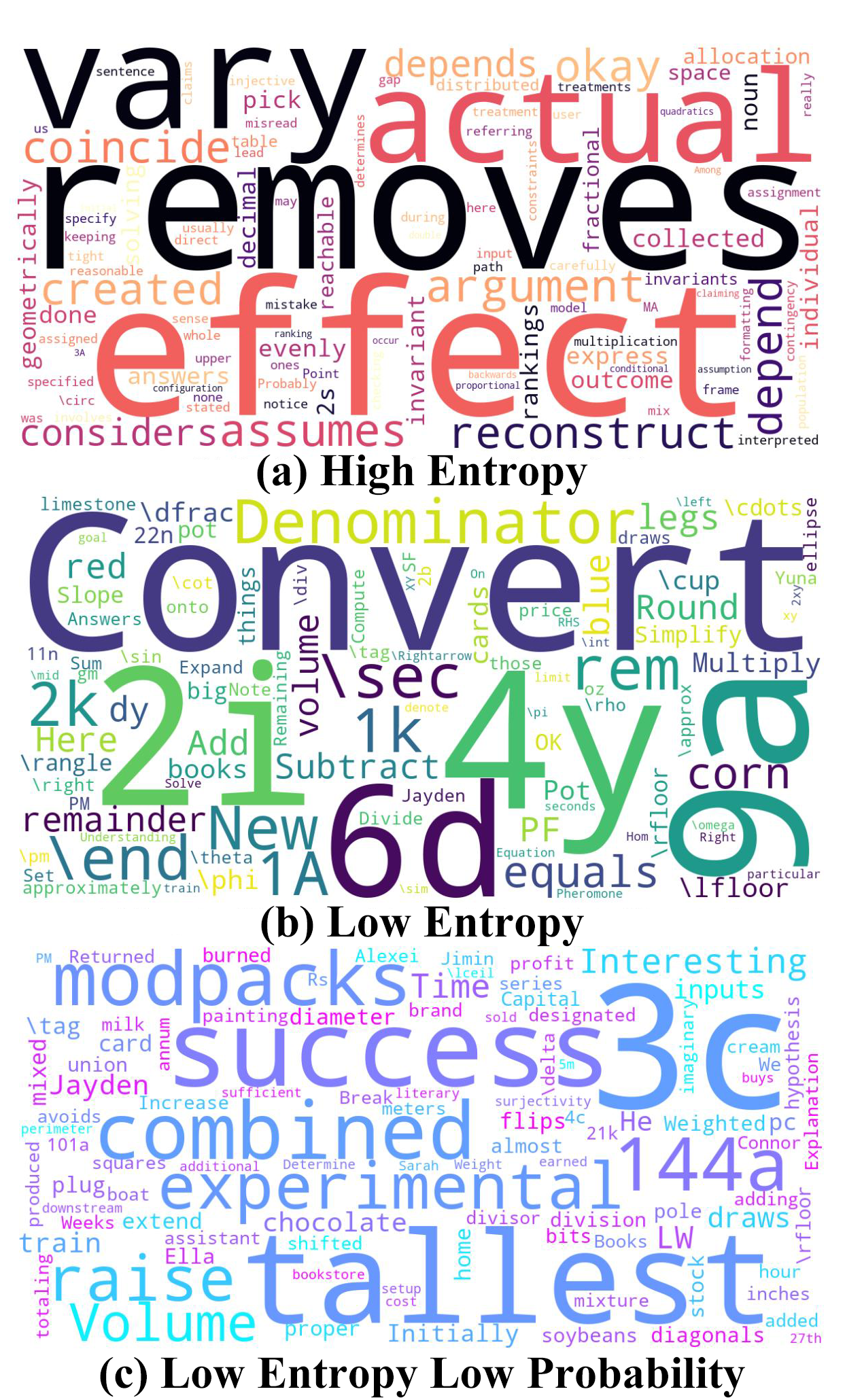}
    \caption{Word clouds visualizing tokens categorized by entropy and probability. \textbf{(a)} High entropy tokens often correspond to reasoning steps. \textbf{(b)} Low entropy tokens correspond to syntax and fixed patterns. \textbf{(c)} \textbf{Confident Conflicts tokens} often involve specific entities or data noise. In these visualizations, \textbf{word size} represents frequency, while \textbf{color} represents entropy.}
    \label{fig:entropy_wordcloud}
\end{figure}

To better understand the motivation behind our Entropy-Adaptive Fine-Tuning loss, we visualize the vocabulary distribution across different entropy and probability regimes in Figure~\ref{fig:entropy_wordcloud}. We categorize tokens into three distinct groups based on the model's predictive uncertainty:

\begin{itemize}[leftmargin=4mm]
    \item \textbf{(a) Branching Points (High Entropy):} As shown in the top panel, high-entropy tokens are predominantly abstract verbs (e.g., \textit{vary, depends, reconstruct}), reasoning connectors, and general nouns. These tokens typically represent semantic \textbf{branching points} where the model faces multiple plausible continuation paths. They often correspond to complex reasoning steps or logical transitions, which are critical for the model to learn during SFT.

    \item \textbf{(b) Confident (Low Entropy):} The middle panel displays tokens where the model exhibits low entropy and high probability. This category is dominated by mathematical symbols (e.g., \textit{Convert, 4y, 6d}), \LaTeX{} syntax (e.g., \texttt{\textbackslash frac}, \texttt{\textbackslash sin}), and functional words. These tokens represent deterministic syntactic structures or rote memorization, which are generally easier for the model to master.

    \item \textbf{(c) Confident Conflicts (Low Entropy and Low probability):} The bottom panel illustrates tokens where the model is confident but potentially incorrect. This group consists largely of specific entities (e.g., \textit{Jayden, modpacks}), rare domain terms, or noise. These often indicate a mismatch between the model distribution and the SFT data, or specific long-tail knowledge.
\end{itemize}

\noindent\textbf{Insight:} This qualitative distinction justifies our entropy-based adaptive strategy. The visualization highlights ``Confident Conflicts'' (Low Entropy, Low Probability) where the model's prior strongly mismatches the SFT data---a primary source of catastrophic forgetting. By differentiating these tokens from critical reasoning steps (High Entropy), our method mitigates the risk of \textbf{destroying the model's original distribution} while effectively adapting to new reasoning patterns.

\section{Open-source Datasets}
\label{app:datasets}
This section details the open-source datasets utilized for training and fine-tuning models, categorized into mathematical reasoning, medical knowledge, and agent capabilities.
\subsection{Mathematical Datasets}
The datasets listed below serve as prompt sources for our data distillation process. We utilized the \textbf{Qwen3-235B-A22B-Instruct-2507~\citep{yang2025qwen3}} model to generate responses to these prompts. For the final training set, we randomly sampled 19k instances where the model's outputs were verified as correct.
\\
\textbf{NuminaMath~\cite{li2024numinamath}:} is a massive dataset dedicated to enhancing the mathematical reasoning capabilities of large language models. Representing the largest public collection in the field, it consists of approximately 860,000 pairs of competition-level math problems and detailed solutions. The dataset covers a broad spectrum of mathematical disciplines and difficulty levels.
\\
\textbf{BigMathVerified~\citep{albalak2025big}:} is a large-scale, high-quality dataset designed specifically for applying reinforcement learning to LMs in the domain of mathematics. It aggregates a vast collection of mathematical problems and solutions from diverse sources, prioritizing data quality and reasoning rigor. By providing a rich environment for training and fine-tuning, Big-Math aims to advance the ability of models to perform reliable multi-step mathematical reasoning and self-correction through reinforcement learning techniques.
\\
\textbf{Nemotron-CrossThink~\citep{akter2025nemotron}:} is a large-scale dataset originally designed to extend self-learning mechanisms across various domains. While the full dataset covers a broad spectrum of reasoning tasks, \textbf{we exclusively select the mathematics subset} to align with our specific research focus. This subset contains high-quality synthetic problems with rigorous logical verification. By isolating these instances, we leverage the dataset's inherent self-correction patterns to specifically enhance the model's precision in math derivation.

\subsection{Medical Datasets}

\textbf{Huatuo-o1~\citep{2412_HuatuoGPT_o1}:} is a specialized supervised fine-tuning (SFT) dataset generated through knowledge distillation from the \textbf{DeepSeek-R1~\citep{guo2025deepseekr1}} model. Built upon the set of verifiable medical problems from \textbf{HuatuoGPT-o1~\citep{2412_HuatuoGPT_o1}}, this dataset leverages DeepSeek-R1 to generate comprehensive, native reasoning chains. The primary objective is to transfer the advanced reasoning capabilities of the R1 model to target models, enabling them to initialize and internalize deep clinical thinking patterns and logical deductions essential for medical tasks.

\subsection{Agent Datasets}
\textbf{Nemotron-Agentic-Tool-Use-v1~\citep{nvidia_nemotron_agentic}} is a synthetic dataset designed to strengthen models' capabilities as interactive, tool-using agents. It focuses on multi-turn conversations where models decompose user goals and execute tool calls to complete tasks. For our experiments, we randomly sampled 20,000 trajectories from the non-thinking subset of this dataset to serve as the training set.

\section{Benchmarks}
\label{app:benchmarks}
In this section, we provide a detailed introduction to the benchmarks used in experiments.

\subsection{General Benchmarks}
\begin{itemize}[leftmargin=1em]
\item \textbf{MMLU~\citep{mmlu}:} is a benchmark in evaluating the massive multitask language understanding and general knowledge capabilities of models. It consists of 57 distinct tasks. All of them are derived from various domains including STEM, the humanities, and the social sciences, ranging from elementary to professional levels. The problems in the MMLU dataset cover a wide variety of subjects such as elementary mathematics, US history, computer science, law, and medicine.

\item \textbf{IFEval~\citep{ifeval}:} is a benchmark designed to assess the instruction-following capabilities of large language models through objective and verifiable means. It consists of approximately 500 prompts containing various verifiable constraints, such as formatting requirements, word count limits, and keyword restrictions. Unlike benchmarks that rely on subjective human or model-based evaluation, IFEval employs a rule-based verification method to determine accuracy. This objective approach allows for a deterministic and reproducible evaluation of a model's ability to strictly adhere to complex user instructions.

\item \textbf{CLUEWSC~\citep{xu2020clue}:} is a benchmark task designed to evaluate coreference resolution and common sense reasoning within the Chinese Language Understanding Evaluation (CLUE) framework. Adapted from the classic Winograd Schema Challenge, the dataset consists of difficult ambiguity resolution problems where a model must identify the correct antecedent of a pronoun in a sentence. The questions are constructed such that changing a single word in the sentence alters the correct reference. Therefore, solving CLUEWSC requires the model to utilize deep contextual understanding and general world knowledge, rather than relying on simple statistical associations or surface-level patterns.

\end{itemize}

\subsection{Mathematical Benchmarks}
\begin{itemize}[leftmargin=1em]
\item \textbf{AIME24~\citep{aime}:} is a dataset in evaluating the mathematical reasoning ability of models. It consists of 30 challenging math problems. All of them are from the American Invitational Mathematics Examination. The problems in the AIME24 dataset cover a wide variety of mathematical fields such as algebraic equations and geometric puzzles. Due to the difficulty characteristics and the richness of question types, it has become a popular benchmark for evaluating the reasoning performance of models, and is widely used in multiple related research experiments.

\item \textbf{AIME25~\citep{aime25}:} consists of 30 challenging math problems. It is directly composed of the real questions from the American Invitational Mathematics Examination (AIME I \& II) newly released in February 2025. AIME25's knowledge areas are extremely wide. It deeply covers core mathematical sections such as algebra, geometry, number theory, and combinatorial mathematics. This characteristic enables the AIME25 dataset to effectively distinguish the mathematical reasoning abilities of different models.

\item \textbf{GSM8K~\citep{GSM8K}:} is an elementary school math problem dataset released by OpenAI. These problems require 2 to 8 steps to solve, mainly through a series of basic calculations to obtain the final answer. This dataset is primarily used to test the logical and mathematical abilities of models and has been applied in multiple benchmark tests.

\end{itemize}

\subsection{Medical Benchmarks}
\begin{itemize}[leftmargin=1em]
\item \textbf{MedMCQA~\citep{pal2022medmcqa}:} is a large-scale multi-subject dataset designed to assess the medical knowledge and reasoning capabilities of models. It comprises over 194,000 multiple-choice questions collected from prestigious Indian medical entrance examinations, such as AIIMS and NEET-PG. The dataset covers a vast spectrum of 21 medical subjects, ranging from basic bio-medical sciences to advanced clinical disciplines like surgery and internal medicine. Given its high difficulty and broad professional coverage, MedMCQA serves as a crucial benchmark for evaluating how well large language models can handle complex healthcare scenarios.
\item \textbf{PubMedQA~\citep{jin2019pubmedqa}:} is a biomedical question answering dataset designed to evaluate reasoning over scientific literature. It is derived from the titles and abstracts of research papers found in the PubMed database. The task requires models to answer research questions with "yes", "no", or "maybe" using the corresponding abstract as context. By focusing on evidence-based reasoning, PubMedQA effectively assesses a model's ability to interpret complex biomedical texts and draw accurate conclusions directly from scientific data.
\item \textbf{MedQA~\citep{medqa}:} is a large-scale dataset specifically designed to evaluate the clinical reasoning and professional medical knowledge of models. The dataset is derived from professional medical board examinations, with its English subset collected from the United States Medical Licensing Examination (USMLE). It consists of complex multiple-choice questions that simulate real-world clinical scenarios, covering diverse topics such as pathology, pharmacology, and patient management. Solving these problems requires deep domain knowledge and the ability to interpret patient case histories. Consequently, MedQA serves as a rigorous benchmark for determining whether models have achieved human-level competency in medicine.
\end{itemize}

\subsection{Agent Benchmarks}
\begin{itemize}[leftmargin=1em]
\item \textbf{BFCL v3~\citep{BFCL}:} is a comprehensive benchmark designed to evaluate the function-calling and tool-use capabilities of large language models. It comprises a diverse set of roughly 2K entries across multiple programming languages, including Python, Java, JavaScript, and REST APIs. The dataset assesses models on complex scenarios ranging from simple function calls to parallel, multiple, and nested calls, as well as multi-turn interactions. By employing an Abstract Syntax Tree (AST) evaluation method, BFCL v3 accurately measures a model's ability to generate syntactically correct and executable API calls, providing a more robust assessment than traditional string-matching metrics.
\end{itemize}

\section{Models}
\label{app:models}
\textbf{Qwen3-4B-Instruct-2507~\citep{yang2025qwen3}:} is a lightweight large language model with 4 billion parameters released by Alibaba Cloud in July 2025. As the instruction-tuned variant of the Qwen3 series, it is engineered for efficient dialogue and general task execution, supporting a native context window of 256K tokens. Unlike its "Thinking" counterpart, this model prioritizes direct response generation without explicit chain-of-thought reasoning, optimizing it for low-latency performance and deployment on consumer-grade hardware.
\\
\textbf{Qwen3-4B-Thinking-2507~\citep{yang2025qwen3}:} is a reasoning lightweight model within the Qwen3 family, featuring 4 billion parameters. Unlike the standard instruction-tuned variant, this model is engineered to perform explicit chain-of-thought (CoT) reasoning, generating intermediate logical steps before producing a final answer. It supports a native context window of 256K tokens and is optimized for complex analytical tasks such as mathematical problem-solving and logical deduction.
\\
\textbf{Qwen2.5-32B-Instruct~\citep{qwen2.5}:} is a 32-billion parameter instruction-tuned large language model released by Alibaba Cloud in September 2024. Positioned as a mid-sized model within the Qwen2.5 series, it is designed to offer an optimal balance between computational efficiency and task performance. The model features a 64-layer Transformer architecture with Grouped Query Attention (GQA) and supports a context window of up to 128K tokens. It demonstrates significant improvements over its predecessors in instruction following, structured data understanding, and logical reasoning, making it highly suitable for deployment in resource-constrained environments that require high-quality generation.
\\
\textbf{GLM-9B-0414~\citep{glm2024chatglm}:} is a lightweight 9-billion parameter language model released by Zhipu AI (THUDM) in April 2025. As an iteration of the GLM-4 open-source series, this model version (0414) is specifically optimized for on-device deployment and agentic tasks. It features a native context window of 128K tokens and demonstrates state-of-the-art performance in tool use (function calling) and long-context understanding among models of similar size. The model is designed to provide a balance between inference latency and task complexity, making it a competitive alternative to larger 32B models for specific downstream applications.
\\
\textbf{Qwen3-235B-A22B-Instruct-2507~\citep{yang2025qwen3}:} is a flagship Mixture-of-Experts (MoE) model released by Alibaba Cloud. While it boasts a massive total parameter count of 235 billion, it utilizes a sparse architecture that activates only 22 billion parameters during inference. This design allows the model to achieve performance comparable to dense state-of-the-art models while maintaining the inference speed and computational efficiency of a much smaller 22B model.

\section{Baselines}
\label{app:baselines}
\begin{itemize}[leftmargin=1em]
\item \textbf{SFT (Supervised Fine-Tuning):} 
    The standard fine-tuning approach that maximizes the likelihood of target tokens using a uniform cross-entropy loss. It treats all tokens equally, making it prone to overfitting specific data patterns and "catastrophic forgetting" of general capabilities.
\item \textbf{SFT$_{KL}$ (SFT with KL Regularization):} 
    A prevalent robust baseline that adds a Kullback-Leibler (KL) divergence penalty to the loss function. It explicitly constrains the policy from deviating too far from the base model (reference model). While effective at preventing forgetting, it incurs significant memory overhead due to the need to maintain a frozen reference model.

 \item \textbf{FLOW \citep{flow}:} 
    A dynamic reweighting method that adjusts the importance of training samples based on their learning dynamics. FLOW monitors the loss trends to identify samples that are likely to cause forgetting, down-weighting them to maintain a smoother optimization trajectory compared to SFT.

\item \textbf{DFT \citep{dft}:} 
\textit{Dynamic Fine-Tuning}. This method reinterprets SFT through a Reinforcement Learning lens, identifying that standard cross-entropy implicitly applies an unstable "inverse-probability weighting" ($1/\pi_\theta$) to gradients. DFT rectifies this by actively scaling the loss with the model's current prediction probability $\pi_\theta(y|x)$. This effectively dampens the gradients for low-probability target tokens (where the model is "surprised" or wrong), stabilizing optimization by preventing aggressive fitting of hard or noisy samples.

    \item \textbf{TALR \citep{talr}:} 
    \textit{Token-Adaptive Loss Reweighting}. A granular approach that assigns varying weights to individual tokens based on their training difficulty (often measured by loss magnitude or gradient norms). TALR aims to focus the model's capacity on "hard" tokens while reducing the impact of easy or noisy tokens, though it typically lacks the uncertainty-awareness of our entropy-based method.
\end{itemize}

\section{Implementation Details}
\label{app:implementation}

In this section, we provide a comprehensive overview of our training infrastructure, hyperparameter configurations, and evaluation protocols to ensure the reproducibility of our results.

\subsection{Training Setup}
All methods were implemented using \textbf{LLaMA-Factory}~\citep{zheng2024llamafactory}, except for SFT$_{KL}$, which was trained using the \textbf{Verl}~\citep{sheng2024hybridflow} framework. All experiments were conducted on 8 NVIDIA A100 GPUs.

\subsection{Hyperparameter Settings}
\label{subsec:hyperparams}

To ensure a fair comparison, we adopted a unified set of hyperparameters across all baseline models and our approach, unless specified otherwise. The specific values for the shared hyperparameters are detailed in Tab.~\ref{tab:hyperparameters}.
\\
Regarding checkpoint selection, we prioritized the checkpoint that demonstrated the best mean performance averaged over all the evaluated benchmarks.

\begin{table}[h]
    \centering
    \begin{tabular}{lc}
    \toprule
    \textbf{Hyperparameter} & \textbf{Value} \\
    \midrule
    Learning Rate & $1 \times 10^{-5}$ \\
    LR Scheduler & Cosine \\
    Warm-up Steps & 0.03 ratio \\
    Optimizer & AdamW \\
    Batch Size & 64 \\
    Num of Epochs & 10 \\
    Max Sequence Length & 16384 \\
    \bottomrule
    \end{tabular}
    \caption{Hyperparameter settings used for all models.}
    \label{tab:hyperparameters}
\end{table}

\paragraph{Method-Specific Hyperparameters.}
The KL-divergence coefficient $\beta$ in SFT$_{KL}$ was set to $0.5$.

\subsection{Evaluation Protocol}
For all evaluations, we conducted three independent runs and reported the average performance.
\\
For mathematical reasoning benchmarks, we strictly followed the evaluation implementation of \textbf{Qwen2.5-Math}~\citep{qwenmath}. For medical domain tasks, we aligned our evaluation pipeline with the \textbf{MedEvalKit} framework~\citep{xu2025lingshu}.

\section{Extended Discussion}
\label{app:discussion}

In this section, we elaborate on the design philosophy of the gating mechanism and position EAFT within the broader landscape of alignment techniques (e.g., RL~\citep{shao2024deepseekmath} and RFT~\citep{touvron2023llama}).

\subsection{Comparison with EAFT$_{sig}$}
\label{subsec:app_sensitivity}

While Section~\ref{subsec:sensitivity} demonstrates that Sigmoid-based gating (\textbf{EAFT$_{sig}$}) achieves competitive performance, we advocate for the linear formulation primarily due to its \textbf{hyperparameter robustness} and \textbf{ease of deployment}.

\paragraph{Sensitivity of Non-Linear Gating.} 
The Sigmoid function introduces two sensitive hyperparameters: steepness ($\alpha$) and the centering threshold ($\beta$). 
\begin{itemize}[leftmargin=4mm]
    \item \textbf{High Sensitivity:} Performance fluctuates significantly with $\beta$. If $\beta$ is too high, the method degenerates into standard SFT; if too low, it mimics a Hard Mask, discarding valuable training signals. This necessitates expensive grid searches for every new dataset or model scale.
    \item \textbf{Binary Bias:} High steepness ($\alpha$) forces a near-binary decision boundary, ignoring uncertainty and treating moderately confident tokens the same as extremely confident ones.
\end{itemize}

\paragraph{The "Parameter-Free" Advantage.} 
The linear formulation acts as a structural prior: \textit{loss weight is directly proportional to uncertainty.} It effectively removes the need for hyperparameter tuning. This "out-of-the-box" robustness ensures EAFT generalizes across domains without requiring the delicate calibration needed for Sigmoid-based variants.

\subsection{Comparison with RL and RFT}
\label{subsec:app_efficiency}

Existing alignment methods like On-policy RL \citep{ppo,shao2024deepseekmath} or Rejection Sampling Fine-Tuning (RFT) \citep{touvron2023llama} mitigate the "alignment tax" but incur high computational or operational costs   \citep{chen2025retaining,rlRazor}. EAFT provides a distinct trade-off favoring efficiency.

\paragraph{Lightweight Efficiency (vs. RL).} 
RL methods typically triple memory requirements by maintaining multiple models simultaneously (Policy, Reference, and Value/Reward models).
\textbf{EAFT} retains the nearly identical memory footprint and computational graph of standard SFT. Crucially, it requires \textbf{no Reference Model} (using the model's own entropy as a proxy for "trustworthiness") and avoids the complex rollout generation phase, significantly accelerating training throughput.

\paragraph{Dynamic Adaptation (vs. RFT).} 
Methods like RFT rely on \textit{static} data curation—filtering or generating data based on the model's capability \textit{before} training begins.
\begin{itemize}[leftmargin=4mm]
    \item \textbf{Static vs. Dynamic:} RFT assumes a fixed knowledge boundary. However, as the model learns, its uncertainty shifts.
    \item \textbf{On-the-fly Correction:} EAFT operates \textit{dynamically}. If the model becomes overconfident about a hallucination mid-training, EAFT automatically down-weights the loss for that specific token. This allows the gating mechanism to evolve in real-time alongside the model parameters, offering an adaptive advantage that static data filtering cannot achieve.
\end{itemize}

\section{LLM Usage}
\label{sec:llm_usage}

We used large language models (LLMs) to improve the clarity and grammatical correctness of the manuscript. After using these tools, the authors carefully reviewed and edited all generated content, and take full responsibility for the final version of the paper.

\setcounter{secnumdepth}{2}
\renewcommand{\thesection}{\Alph{section}}
\renewcommand{\thesubsection}{\Alph{section}.\arabic{subsection}}


\begin{thebibliography}{48}
\providecommand{\natexlab}[1]{#1}

\bibitem[{AI-MO(2025)}]{aime}
AI-MO. 2025.
\newblock Ai-mo/aimo-validation-aime · datasets at hugging face.
\newblock \url{https://huggingface.co/datasets/AI-MO/aimo-validation-aime}.
\newblock Online; accessed 2025-12-28.

\bibitem[{Akter et~al.(2025)Akter, Prabhumoye, Novikov, Han, Lin, Bakhturina, Nyberg, Choi, Patwary, Shoeybi et~al.}]{akter2025nemotron}
Syeda~Nahida Akter, Shrimai Prabhumoye, Matvei Novikov, Seungju Han, Ying Lin, Evelina Bakhturina, Eric Nyberg, Yejin Choi, Mostofa Patwary, Mohammad Shoeybi, and 1 others. 2025.
\newblock Nemotron-crossthink: Scaling self-learning beyond math reasoning.
\newblock \emph{arXiv preprint arXiv:2504.13941}.

\bibitem[{Albalak et~al.(2025)Albalak, Phung, Lile, Rafailov, Gandhi, Castricato, Singh, Blagden, Xiang, Mahan et~al.}]{albalak2025big}
Alon Albalak, Duy Phung, Nathan Lile, Rafael Rafailov, Kanishk Gandhi, Louis Castricato, Anikait Singh, Chase Blagden, Violet Xiang, Dakota Mahan, and 1 others. 2025.
\newblock Big-math: A large-scale, high-quality math dataset for reinforcement learning in language models.
\newblock \emph{arXiv preprint arXiv:2502.17387}.

\bibitem[{Askell et~al.(2021)Askell, Bai, Chen, Drain, Ganguli, Henighan, Jones, Joseph, Mann, DasSarma et~al.}]{askell2021general}
Amanda Askell, Yuntao Bai, Anna Chen, Dawn Drain, Deep Ganguli, Tom Henighan, Andy Jones, Nicholas Joseph, Ben Mann, Nova DasSarma, and 1 others. 2021.
\newblock A general language assistant as a laboratory for alignment.
\newblock \emph{arXiv preprint arXiv:2112.00861}.

\bibitem[{Chen et~al.(2025)Chen, Razin, Narasimhan, and Chen}]{chen2025retaining}
Howard Chen, Noam Razin, Karthik Narasimhan, and Danqi Chen. 2025.
\newblock Retaining by doing: The role of on-policy data in mitigating forgetting.
\newblock \emph{arXiv preprint arXiv:2510.18874}.

\bibitem[{Chen et~al.(2024)Chen, Cai, Ji, Wang, Liu, Wang, Hou, and Wang}]{2412_HuatuoGPT_o1}
Junying Chen, Zhenyang Cai, Ke~Ji, Xidong Wang, Wanlong Liu, Rongsheng Wang, Jianye Hou, and Benyou Wang. 2024.
\newblock Huatuogpt-o1, towards medical complex reasoning with llms.
\newblock \emph{arXiv preprint arXiv:2412.18925}.

\bibitem[{Chu et~al.(2025)Chu, Zhai, Yang, Tong, Xie, Schuurmans, Le, Levine, and Ma}]{chu2025sftmemorizes}
Tianzhe Chu, Yuexiang Zhai, Jihan Yang, Shengbang Tong, Saining Xie, Dale Schuurmans, Quoc~V Le, Sergey Levine, and Yi~Ma. 2025.
\newblock Sft memorizes, rl generalizes: A comparative study of foundation model post-training.
\newblock \emph{arXiv preprint arXiv:2501.17161}.

\bibitem[{Cobbe et~al.(2021)Cobbe, Kosaraju, Bavarian, Chen, Jun, Kaiser, Plappert, Tworek, Hilton, Nakano et~al.}]{GSM8K}
Karl Cobbe, Vineet Kosaraju, Mohammad Bavarian, Mark Chen, Heewoo Jun, Lukasz Kaiser, Matthias Plappert, Jerry Tworek, Jacob Hilton, Reiichiro Nakano, and 1 others. 2021.
\newblock Training verifiers to solve math word problems.
\newblock \emph{arXiv preprint arXiv:2110.14168}.

\bibitem[{Fan et~al.(2018)Fan, Lewis, and Dauphin}]{fan2018hierarchical}
Angela Fan, Mike Lewis, and Yann Dauphin. 2018.
\newblock Hierarchical neural story generation.
\newblock \emph{arXiv preprint arXiv:1805.04833}.

\bibitem[{GLM et~al.(2024)GLM, Zeng, Xu, Wang, Zhang, Yin, Rojas, Feng, Zhao, Lai, Yu, Wang, Sun, Zhang, Cheng, Gui, Tang, Zhang, Li, Zhao, Wu, Zhong, Liu, Huang, Zhang, Zheng, Lu, Duan, Zhang, Cao, Yang, Tam, Zhao, Liu, Xia, Zhang, Gu, Lv, Liu, Liu, Yang, Song, Zhang, An, Xu, Niu, Yang, Li, Bai, Dong, Qi, Wang, Yang, Du, Hou, and Wang}]{glm2024chatglm}
Team GLM, Aohan Zeng, Bin Xu, Bowen Wang, Chenhui Zhang, Da~Yin, Diego Rojas, Guanyu Feng, Hanlin Zhao, Hanyu Lai, Hao Yu, Hongning Wang, Jiadai Sun, Jiajie Zhang, Jiale Cheng, Jiayi Gui, Jie Tang, Jing Zhang, Juanzi Li, and 37 others. 2024.
\newblock \href {https://arxiv.org/abs/2406.12793} {Chatglm: A family of large language models from glm-130b to glm-4 all tools}.
\newblock \emph{Preprint}, arXiv:2406.12793.

\bibitem[{Guo et~al.(2025)Guo, Yang, Zhang, Song, Wang, Zhu, Xu, Zhang, Ma, Bi et~al.}]{guo2025deepseekr1}
Daya Guo, Dejian Yang, Haowei Zhang, Junxiao Song, Peiyi Wang, Qihao Zhu, Runxin Xu, Ruoyu Zhang, Shirong Ma, Xiao Bi, and 1 others. 2025.
\newblock Deepseek-r1 incentivizes reasoning in llms through reinforcement learning.
\newblock \emph{Nature}, 645(8081):633--638.

\bibitem[{Hendrycks et~al.(2020)Hendrycks, Burns, Basart, Zou, Mazeika, Song, and Steinhardt}]{mmlu}
Dan Hendrycks, Collin Burns, Steven Basart, Andy Zou, Mantas Mazeika, Dawn Song, and Jacob Steinhardt. 2020.
\newblock Measuring massive multitask language understanding.
\newblock \emph{arXiv preprint arXiv:2009.03300}.

\bibitem[{Holtzman et~al.(2019)Holtzman, Buys, Du, Forbes, and Choi}]{holtzman2019curious}
Ari Holtzman, Jan Buys, Li~Du, Maxwell Forbes, and Yejin Choi. 2019.
\newblock The curious case of neural text degeneration.
\newblock \emph{arXiv preprint arXiv:1904.09751}.

\bibitem[{Jin et~al.(2021)Jin, Pan, Oufattole, Weng, Fang, and Szolovits}]{medqa}
Di~Jin, Eileen Pan, Nassim Oufattole, Wei-Hung Weng, Hanyi Fang, and Peter Szolovits. 2021.
\newblock What disease does this patient have? a large-scale open domain question answering dataset from medical exams.
\newblock \emph{Applied Sciences}, 11(14):6421.

\bibitem[{Jin et~al.(2019)Jin, Dhingra, Liu, Cohen, and Lu}]{jin2019pubmedqa}
Qiao Jin, Bhuwan Dhingra, Zhengping Liu, William Cohen, and Xinghua Lu. 2019.
\newblock Pubmedqa: A dataset for biomedical research question answering.
\newblock In \emph{Proceedings of the 2019 conference on empirical methods in natural language processing and the 9th international joint conference on natural language processing (EMNLP-IJCNLP)}, pages 2567--2577.

\bibitem[{Kirkpatrick et~al.(2017)Kirkpatrick, Pascanu, Rabinowitz, Veness, Desjardins, Rusu, Milan, Quan, Ramalho, Grabska-Barwinska et~al.}]{kirkpatrick2017overcoming}
James Kirkpatrick, Razvan Pascanu, Neil Rabinowitz, Joel Veness, Guillaume Desjardins, Andrei~A Rusu, Kieran Milan, John Quan, Tiago Ramalho, Agnieszka Grabska-Barwinska, and 1 others. 2017.
\newblock Overcoming catastrophic forgetting in neural networks.
\newblock \emph{Proceedings of the national academy of sciences}, 114(13):3521--3526.

\bibitem[{Li et~al.(2024)Li, Beeching, Tunstall, Lipkin, Soletskyi, Huang, Rasul, Yu, Jiang, Shen et~al.}]{li2024numinamath}
Jia Li, Edward Beeching, Lewis Tunstall, Ben Lipkin, Roman Soletskyi, Shengyi Huang, Kashif Rasul, Longhui Yu, Albert~Q Jiang, Ziju Shen, and 1 others. 2024.
\newblock Numinamath: The largest public dataset in ai4maths with 860k pairs of competition math problems and solutions.
\newblock \emph{Hugging Face repository}, 13(9):9.

\bibitem[{Li and Hoiem(2017)}]{li2017learning}
Zhizhong Li and Derek Hoiem. 2017.
\newblock Learning without forgetting.
\newblock \emph{IEEE transactions on pattern analysis and machine intelligence}, 40(12):2935--2947.

\bibitem[{Lin et~al.(2025)Lin, Wang, Qian, Wang, Srinivasan, Zeng, Jiao, Zhou, Gesi, Wang et~al.}]{talr}
Jiacheng Lin, Zhongruo Wang, Kun Qian, Tian Wang, Arvind Srinivasan, Hansi Zeng, Ruochen Jiao, Xie Zhou, Jiri Gesi, Dakuo Wang, and 1 others. 2025.
\newblock Sft doesn't always hurt general capabilities: Revisiting domain-specific fine-tuning in llms.
\newblock \emph{arXiv preprint arXiv:2509.20758}.

\bibitem[{Lopez-Paz and Ranzato(2017)}]{lopez2017gradient}
David Lopez-Paz and Marc'Aurelio Ranzato. 2017.
\newblock Gradient episodic memory for continual learning.
\newblock \emph{Advances in neural information processing systems}, 30.

\bibitem[{Luo et~al.(2023)Luo, Yang, Meng, Li, Zhou, and Zhang}]{luo2023empirical}
Yun Luo, Zhen Yang, Fandong Meng, Yafu Li, Jie Zhou, and Yue Zhang. 2023.
\newblock An empirical study of catastrophic forgetting in large language models during continual fine-tuning.
\newblock \emph{arXiv e-prints}, pages arXiv--2308.

\bibitem[{McCloskey and Cohen(1989)}]{mccloskey1989catastrophic}
Michael McCloskey and Neal~J Cohen. 1989.
\newblock Catastrophic interference in connectionist networks: The sequential learning problem.
\newblock In \emph{Psychology of learning and motivation}, volume~24, pages 109--165. Elsevier.

\bibitem[{Mukherjee et~al.(2025)Mukherjee, Yuan, Hakkani-Tur, and Peng}]{mukherjee2025reinforcement}
Sagnik Mukherjee, Lifan Yuan, Dilek Hakkani-Tur, and Hao Peng. 2025.
\newblock Reinforcement learning finetunes small subnetworks in large language models.
\newblock \emph{arXiv preprint arXiv:2505.11711}.

\bibitem[{{NVIDIA}(2025)}]{nvidia_nemotron_agentic}
{NVIDIA}. 2025.
\newblock \href {https://huggingface.co/datasets/nvidia/Nemotron-Agentic-v1} {Nemotron-agentic-v1 dataset}.
\newblock Hugging Face.
\newblock Accessed: 2026-01-04.

\bibitem[{Ouyang et~al.(2022)Ouyang, Wu, Jiang, Almeida, Wainwright, Mishkin, Zhang, Agarwal, Slama, Ray et~al.}]{ouyang2022training}
Long Ouyang, Jeffrey Wu, Xu~Jiang, Diogo Almeida, Carroll Wainwright, Pamela Mishkin, Chong Zhang, Sandhini Agarwal, Katarina Slama, Alex Ray, and 1 others. 2022.
\newblock Training language models to follow instructions with human feedback.
\newblock \emph{Advances in neural information processing systems}, 35:27730--27744.

\bibitem[{Pal et~al.(2022)Pal, Umapathi, and Sankarasubbu}]{pal2022medmcqa}
Ankit Pal, Logesh~Kumar Umapathi, and Malaikannan Sankarasubbu. 2022.
\newblock Medmcqa: A large-scale multi-subject multi-choice dataset for medical domain question answering.
\newblock In \emph{Conference on health, inference, and learning}, pages 248--260. PMLR.

\bibitem[{Patil et~al.()Patil, Mao, Yan, Ji, Suresh, Stoica, and Gonzalez}]{BFCL}
Shishir~G Patil, Huanzhi Mao, Fanjia Yan, Charlie Cheng-Jie Ji, Vishnu Suresh, Ion Stoica, and Joseph~E Gonzalez.
\newblock The berkeley function calling leaderboard (bfcl): From tool use to agentic evaluation of large language models.
\newblock In \emph{Forty-second International Conference on Machine Learning}.

\bibitem[{Qwen et~al.(2024)Qwen, :, Yang, Yang, Zhang, Hui, Zheng, Yu, Li, Liu, Huang, Wei, Lin, Yang, Tu, Zhang, Yang, Yang, Zhou, Lin, Dang, Lu, Bao, Yang, Yu, Li, Xue, Zhang, Zhu, Men, Lin, Li, Xia, Ren, Ren, Fan, Su, Zhang, Wan, Liu, Cui, Zhang, and Qiu}]{qwen2.5}
Qwen, :, An~Yang, Baosong Yang, Beichen Zhang, Binyuan Hui, Bo~Zheng, Bowen Yu, Chengyuan Li, Dayiheng Liu, Fei Huang, Haoran Wei, Huan Lin, Jian Yang, Jianhong Tu, Jianwei Zhang, Jianxin Yang, Jiaxi Yang, Jingren Zhou, and 24 others. 2024.
\newblock \href {https://arxiv.org/abs/2412.15115} {Qwen2.5 technical report}.
\newblock \emph{Preprint}, arXiv:2412.15115.

\bibitem[{Radford et~al.(2019)Radford, Wu, Child, Luan, Amodei, Sutskever et~al.}]{radford2019language}
Alec Radford, Jeffrey Wu, Rewon Child, David Luan, Dario Amodei, Ilya Sutskever, and 1 others. 2019.
\newblock Language models are unsupervised multitask learners.
\newblock \emph{OpenAI blog}, 1(8):9.

\bibitem[{Razin et~al.(2023)Razin, Zhou, Saremi, Thilak, Bradley, Nakkiran, Susskind, and Littwin}]{razin2023vanishing}
Noam Razin, Hattie Zhou, Omid Saremi, Vimal Thilak, Arwen Bradley, Preetum Nakkiran, Joshua Susskind, and Etai Littwin. 2023.
\newblock Vanishing gradients in reinforcement finetuning of language models.
\newblock \emph{arXiv preprint arXiv:2310.20703}.

\bibitem[{Sanyal et~al.(2025)Sanyal, Prairie, Das, Kavis, and Sanghavi}]{flow}
Sunny Sanyal, Hayden Prairie, Rudrajit Das, Ali Kavis, and Sujay Sanghavi. 2025.
\newblock Upweighting easy samples in fine-tuning mitigates forgetting.
\newblock \emph{arXiv preprint arXiv:2502.02797}.

\bibitem[{Schulman et~al.(2017)Schulman, Wolski, Dhariwal, Radford, and Klimov}]{ppo}
John Schulman, Filip Wolski, Prafulla Dhariwal, Alec Radford, and Oleg Klimov. 2017.
\newblock Proximal policy optimization algorithms.
\newblock \emph{arXiv preprint arXiv:1707.06347}.

\bibitem[{Shao et~al.(2024)Shao, Wang, Zhu, Xu, Song, Bi, Zhang, Zhang, Li, Wu et~al.}]{shao2024deepseekmath}
Zhihong Shao, Peiyi Wang, Qihao Zhu, Runxin Xu, Junxiao Song, Xiao Bi, Haowei Zhang, Mingchuan Zhang, YK~Li, Yang Wu, and 1 others. 2024.
\newblock Deepseekmath: Pushing the limits of mathematical reasoning in open language models.
\newblock \emph{arXiv preprint arXiv:2402.03300}.

\bibitem[{Shenfeld et~al.(2025)Shenfeld, Pari, and Agrawal}]{rlRazor}
Idan Shenfeld, Jyothish Pari, and Pulkit Agrawal. 2025.
\newblock Rl's razor: Why online reinforcement learning forgets less.
\newblock \emph{arXiv preprint arXiv:2509.04259}.

\bibitem[{Sheng et~al.(2024)Sheng, Zhang, Ye, Wu, Zhang, Zhang, Peng, Lin, and Wu}]{sheng2024hybridflow}
Guangming Sheng, Chi Zhang, Zilingfeng Ye, Xibin Wu, Wang Zhang, Ru~Zhang, Yanghua Peng, Haibin Lin, and Chuan Wu. 2024.
\newblock Hybridflow: A flexible and efficient rlhf framework.
\newblock \emph{arXiv preprint arXiv: 2409.19256}.

\bibitem[{Shi et~al.(2025)Shi, Xu, Wang, Qin, Wang, Wang, Wang, Ebrahimi, and Wang}]{shi2025continual}
Haizhou Shi, Zihao Xu, Hengyi Wang, Weiyi Qin, Wenyuan Wang, Yibin Wang, Zifeng Wang, Sayna Ebrahimi, and Hao Wang. 2025.
\newblock Continual learning of large language models: A comprehensive survey.
\newblock \emph{ACM Computing Surveys}, 58(5):1--42.

\bibitem[{Team et~al.(2025)Team, Bai, Bao, Chen, Chen, Chen, Chen, Chen, Chen, Chen et~al.}]{k2}
Kimi Team, Yifan Bai, Yiping Bao, Guanduo Chen, Jiahao Chen, Ningxin Chen, Ruijue Chen, Yanru Chen, Yuankun Chen, Yutian Chen, and 1 others. 2025.
\newblock Kimi k2: Open agentic intelligence.
\newblock \emph{arXiv preprint arXiv:2507.20534}.

\bibitem[{Touvron et~al.(2023)Touvron, Martin, Stone, Albert, Almahairi, Babaei, Bashlykov, Batra, Bhargava, Bhosale et~al.}]{touvron2023llama}
Hugo Touvron, Louis Martin, Kevin Stone, Peter Albert, Amjad Almahairi, Yasmine Babaei, Nikolay Bashlykov, Soumya Batra, Prajjwal Bhargava, Shruti Bhosale, and 1 others. 2023.
\newblock Llama 2: Open foundation and fine-tuned chat models.
\newblock \emph{arXiv preprint arXiv:2307.09288}.

\bibitem[{Wang et~al.(2025)Wang, Cheng, Peng, Bao, Li, Guo, Li, Zeng, Zhou, and Qiu}]{wang2025implicit}
Bo~Wang, Qinyuan Cheng, Runyu Peng, Rong Bao, Peiji Li, Qipeng Guo, Linyang Li, Zhiyuan Zeng, Yunhua Zhou, and Xipeng Qiu. 2025.
\newblock Implicit reward as the bridge: A unified view of sft and dpo connections.
\newblock \emph{arXiv preprint arXiv:2507.00018}.

\bibitem[{Wu et~al.(2025)Wu, Zhou, Ziheng, Peng, Ye, Hu, Zhu, Qi, Yang, and Yang}]{dft}
Yongliang Wu, Yizhou Zhou, Zhou Ziheng, Yingzhe Peng, Xinyu Ye, Xinting Hu, Wenbo Zhu, Lu~Qi, Ming-Hsuan Yang, and Xu~Yang. 2025.
\newblock On the generalization of sft: A reinforcement learning perspective with reward rectification.
\newblock \emph{arXiv preprint arXiv:2508.05629}.

\bibitem[{Xu et~al.(2020)Xu, Hu, Zhang, Li, Cao, Li, Yxn, Bai, Shu, and Xi}]{xu2020clue}
Liang Xu, Hai Hu, Xuanwei Zhang, Lu~Li, Chenjie Cao, Yudong Li, Yechen Yxn, Shushan Bai, Man Shu, and Xiangang Xi. 2020.
\newblock \href {https://aclanthology.org/2020.coling-main.419} {{CLUE}: A {C}hinese {L}anguage {U}nderstanding {E}valuation {B}enchmark}.
\newblock In \emph{Proceedings of the 28th International Conference on Computational Linguistics}, pages 4762--4772.

\bibitem[{Xu et~al.(2025)Xu, Chan, Li, Aljunied, Yuan, Wang, Xiao, Chen, Liu, Li et~al.}]{xu2025lingshu}
Weiwen Xu, Hou~Pong Chan, Long Li, Mahani Aljunied, Ruifeng Yuan, Jianyu Wang, Chenghao Xiao, Guizhen Chen, Chaoqun Liu, Zhaodonghui Li, and 1 others. 2025.
\newblock Lingshu: A generalist foundation model for unified multimodal medical understanding and reasoning.
\newblock \emph{arXiv preprint arXiv:2506.07044}.

\bibitem[{Yang et~al.(2025)Yang, Li, Yang, Zhang, Hui, Zheng, Yu, Gao, Huang, Lv et~al.}]{yang2025qwen3}
An~Yang, Anfeng Li, Baosong Yang, Beichen Zhang, Binyuan Hui, Bo~Zheng, Bowen Yu, Chang Gao, Chengen Huang, Chenxu Lv, and 1 others. 2025.
\newblock Qwen3 technical report.
\newblock \emph{arXiv preprint arXiv:2505.09388}.

\bibitem[{Yang et~al.(2024)Yang, Zhang, Hui, Gao, Yu, Li, Liu, Tu, Zhou, Lin et~al.}]{qwenmath}
An~Yang, Beichen Zhang, Binyuan Hui, Bofei Gao, Bowen Yu, Chengpeng Li, Dayiheng Liu, Jianhong Tu, Jingren Zhou, Junyang Lin, and 1 others. 2024.
\newblock Qwen2. 5-math technical report: Toward mathematical expert model via self-improvement.
\newblock \emph{arXiv preprint arXiv:2409.12122}.

\bibitem[{Zeng et~al.(2025)Zeng, Huang, Liu, Liu, He, Ma, and He}]{zeng2025simplerl}
Weihao Zeng, Yuzhen Huang, Qian Liu, Wei Liu, Keqing He, Zejun Ma, and Junxian He. 2025.
\newblock Simplerl-zoo: Investigating and taming zero reinforcement learning for open base models in the wild.
\newblock \emph{arXiv preprint arXiv:2503.18892}.

\bibitem[{Zhang and Math-AI(2025)}]{aime25}
Yifan Zhang and Team Math-AI. 2025.
\newblock American invitational mathematics examination (aime) 2025.

\bibitem[{Zheng et~al.(2024)Zheng, Zhang, Zhang, Ye, Luo, Feng, and Ma}]{zheng2024llamafactory}
Yaowei Zheng, Richong Zhang, Junhao Zhang, Yanhan Ye, Zheyan Luo, Zhangchi Feng, and Yongqiang Ma. 2024.
\newblock \href {http://arxiv.org/abs/2403.13372} {Llamafactory: Unified efficient fine-tuning of 100+ language models}.
\newblock In \emph{Proceedings of the 62nd Annual Meeting of the Association for Computational Linguistics (Volume 3: System Demonstrations)}, Bangkok, Thailand. Association for Computational Linguistics.

\bibitem[{Zhou et~al.(2023)Zhou, Lu, Mishra, Brahma, Basu, Luan, Zhou, and Hou}]{ifeval}
Jeffrey Zhou, Tianjian Lu, Swaroop Mishra, Siddhartha Brahma, Sujoy Basu, Yi~Luan, Denny Zhou, and Le~Hou. 2023.
\newblock Instruction-following evaluation for large language models.
\newblock \emph{arXiv preprint arXiv:2311.07911}.

\end{thebibliography}
\end{document}